\documentclass[sn-mathphys,Numbered]{sn-jnl}

\usepackage{graphicx}%
\usepackage{subfigure}
\usepackage{multirow}%
\usepackage{amsmath,amssymb,amsfonts}%
\usepackage{amsthm}%
\usepackage{mathrsfs}%
\usepackage[title]{appendix}%
\usepackage{xcolor}%
\usepackage{textcomp}%
\usepackage{manyfoot}%
\usepackage{booktabs}%
\usepackage{algorithm}%
\usepackage{algorithmicx}%
\usepackage{algpseudocode}%
\usepackage{listings}%
\usepackage{ragged2e}

\definecolor{rebuttle}{rgb}{0, 0, 0}
\newcommand{\op}[1]{\operatorname{#1}}
\newcommand{\mbf}[1]{\mathbf{#1}}

\theoremstyle{thmstyleone}%
\theoremstyle{thmstyletwo}%

\theoremstyle{thmstylethree}%

\begin{document}
	
	\title{Human-imperceptible, Machine-recognizable Images}
	
	
	\author[1,2]{\fnm{Fusheng} \sur{Hao}}\email{fs.hao@siat.ac.cn}
	
	\author[3,4]{\fnm{Fengxiang} \sur{He}}\email{F.He@ed.ac.uk}
	
	\author[5]{\fnm{Yikai} \sur{Wang}}\email{yikaiw@outlook.com}
	
	\author[1,2]{\fnm{Fuxiang} \sur{Wu}}\email{fx.wu1@siat.ac.cn}
	
	\author[6]{\fnm{Jing} \sur{Zhang}}\email{jing.zhang1@sydney.edu.au}
	
	\author[1,2]{\fnm{Jun} \sur{Cheng}}\email{jun.cheng@siat.ac.cn}
	
	\author[6,4]{\fnm{Dacheng} \sur{Tao}}\email{dacheng.tao@sydney.edu.au}
	
	\affil[1]{\orgname{Shenzhen Institute of Advanced Technology, CAS}
	}
	
	\affil[2]{
		\orgname{The Chinese University of Hong Kong}
	}
	
	\affil[3]{\orgdiv{AIAI, School of Informatics}, \orgname{University of Edinburgh}
	}
	
	\affil[4]{\orgdiv{JD Explore Academy}, \orgname{JD.com, Inc}
	}
	
	\affil[5]{\orgdiv{Deptment of Computer Science and Technology}, 
		\orgname{Tsinghua University}
	}
	
	\affil[6]{\orgdiv{School of Computer Science}, \orgname{The University of Sydney}
	}
	

	\abstract{
		Massive human-related data is collected to train neural networks for computer vision tasks. A major conflict is exposed relating to software engineers between better developing AI systems and distancing from the sensitive training data. To reconcile this conflict, this paper proposes an efficient privacy-preserving learning paradigm, where images are first encrypted to become ``human-imperceptible, machine-recognizable'' via one of the two encryption strategies: (1) random shuffling to a set of equally-sized patches and (2) mixing-up sub-patches of the images. Then, minimal adaptations are made to vision transformer to enable it to learn on the encrypted images for vision tasks, including image classification and object detection. Extensive experiments on ImageNet and COCO show that the proposed paradigm achieves comparable accuracy with the competitive methods. Decrypting the encrypted images requires solving an NP-hard jigsaw puzzle or an ill-posed inverse problem, which is empirically shown intractable to be recovered by various attackers, including the powerful vision transformer-based attacker. We thus show that the proposed paradigm can ensure the encrypted images have become human-imperceptible while preserving machine-recognizable information. The code is available at \url{https://github.com/FushengHao/PrivacyPreservingML.}}
	
	\maketitle
	
	\section{Introduction}
	Relying on massive personal images, the industry has shown promising capabilities for developing artificial intelligence (AI) for many computer vision tasks, e.g., image classification~\cite{he2016deep, ijcv3}, face recognition~\cite{li2021spherical, ijcv4}, action recognition~\cite{ijcv5, ijcv6}, etc. In this process, a major conflict has been seen relating to software engineers between better developing AI systems and distancing from the sensitive training data. To reconcile this conflict, we raise a problem in this paper:
	\bigskip
	\begin{center}
		{\it Can we process images to be human-imperceptible and machine-recognizable?}
	\end{center}
	\bigskip
	\noindent Such that software engineers can access the sensitive training data to facilitate their developing AI systems, without leaking the sensitive contents to the engineers.
	
\begin{figure*}[t]
		\centering
		\includegraphics[width=0.95\linewidth]{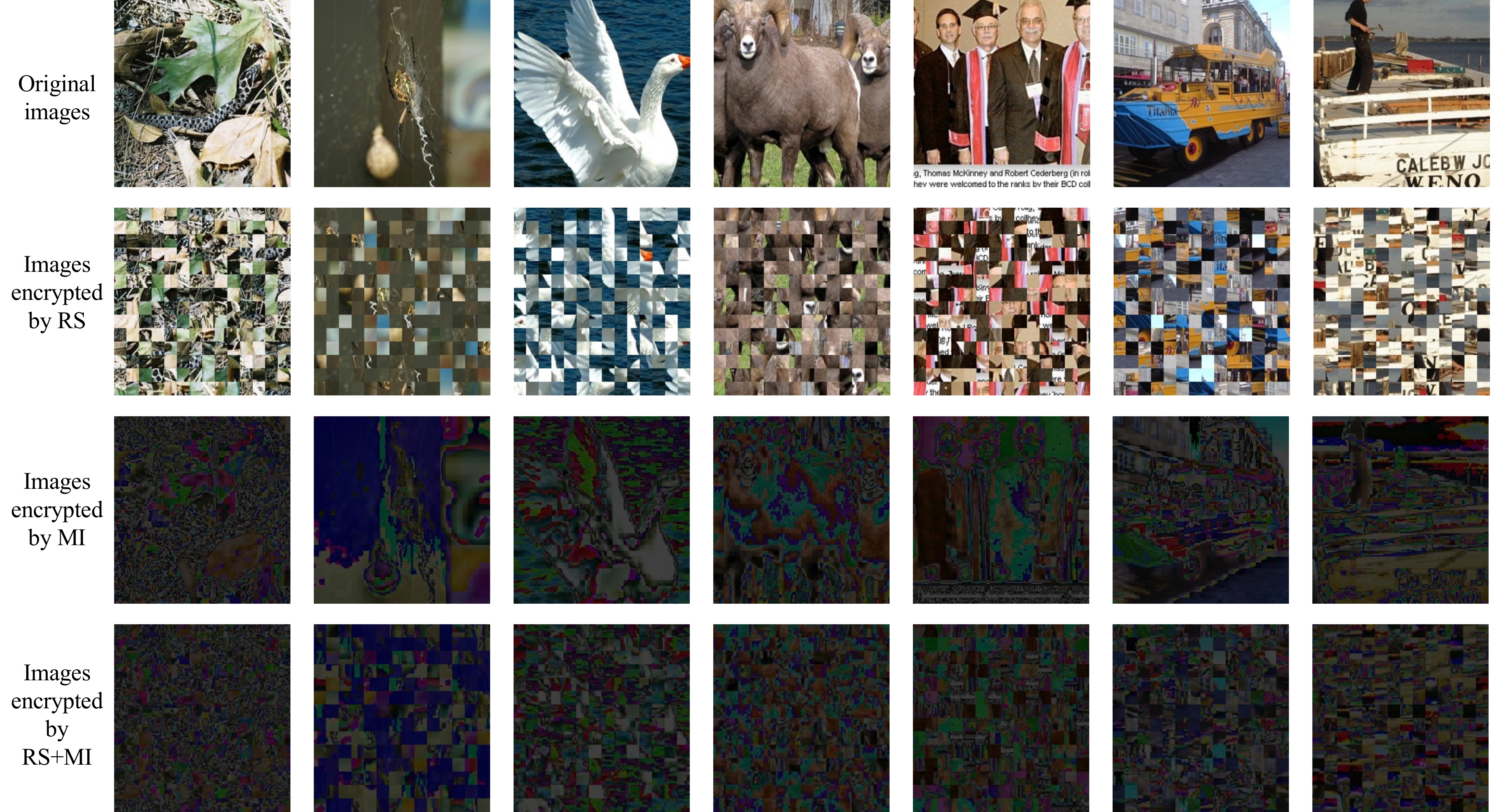}
		\vspace{-0.0em}
		\caption{Illustration of images encrypted by RS, MI, and their combination. The visual contents of encrypted images are near-completely protected from recognizing by human eyes.}
		\label{fig:encrypted_images}
	\end{figure*}

	To process images to be ``human-imperceptible, machine-recognizable'', we develop two encryption strategies: random shuffling (RS) and mixing-up (MI); see Figure~\ref{fig:encrypted_images}. RS randomly shuffles the patch order of an image, which destroys the position configurations between patches and can thus be applied to position-insensitive scenarios. Decrypting an image encrypted by RS is to solve a jigsaw puzzle problem, which can incur considerable computational overhead since the problem to be solved is a NP-hard one~\cite{Demaine}. MI mixes up the sub-patches in a patch, which preserves the position configurations between patches and can thus be applied to position-sensitive scenarios. Decrypting an image encrypted by MI is to solve an ill-posed inverse problem, which could be hard to solve due to the difficulty in modeling the sub-patch distribution.
	
\begin{figure*}[t]
		\centering
		\includegraphics[width=0.90\linewidth]{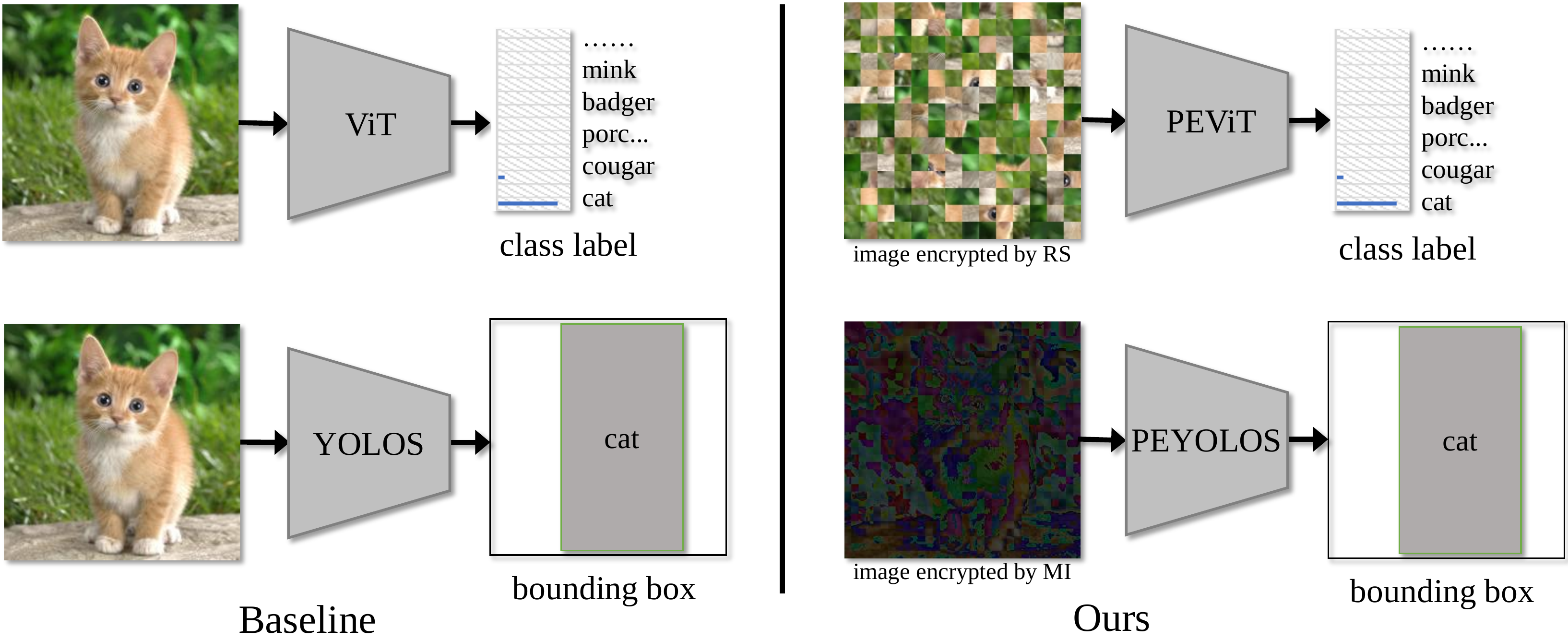}
		\vspace{-0.0em}
		\caption{Images encrypted by RS and MI are still machine-learnable, by further designing architectures PEViT and PEYOLOS, based on ViT and YOLOS.}
		\label{fig:vs}
	\end{figure*}

	Then, we make minimal adaptations to vision transformer (ViT)~\cite{dosovitskiy2020image} to enable it to learn on the encrypted images; see Figure~\ref{fig:vs}. By removing position encoding, ViT is made permutation-invariant and thus capable of learning on images encrypted by RS, resulting in PEViT for the image classification task. Further, we develop a reference-based positional encoding for PEViT, which can retain the permutation-invariant property and thus boost the performance by a noticeable margin. Since position information plays a key role in the low-level object detection task, MI is chosen to encrypt images for this task. By adapting the way that image patches are embedded, YOLOS~\cite{YOLOS}, a vanilla ViT-based model, is able to learn on images encrypted by MI, resulting in PEViT for the object detection task. 
	
	We conduct extensive experiments on ImageNet~\cite{deng2009imagenet} and COCO~\cite{lin2014microsoft}. Extensive attack experiments show the security of our encryption strategies. Comparison results on large-scale benchmarks show that both PEViT and PEYOLOS achieve promising performance even with highly encrypted images as input. We thus show that the proposed paradigm can ensure the encrypted images have become human-imperceptible while preserving machine-recognizable information. The code is available at \url{https://github.com/FushengHao/PrivacyPreservingML.}

	Our main contributions are summarized as follows:
	
	\begin{itemize}
		\item 
		We propose an efficient privacy-preserving learning paradigm that can ensure the encrypted images have become human-imperceptible while preserving machine-recognizable information.
		
		\item 
		RS is tailored for the standard image classification with ViT. By substituting the reference-based positional encoding for the original one, ViT is capable of learning on images encrypted by RS.
		
		\item 
		By further designing MI, the privacy-preserving learning paradigm is extensible to position-sensitive tasks, such as object detection, for which we only need to adapt the way image patches are embedded.
		
		\item 
		Extensive experiments demonstrate the effectiveness of the proposed privacy-preserving learning paradigm.
		
		
		
	\end{itemize}
	
	\section{Related Work}
	
	\textbf{Vision transformers.} 
	Self-attention based Transformer~\cite{vaswani2017attention} has achieved great success in natural language processing. To make Transformer suitable for image classification, the pioneering work of ViT~\cite{dosovitskiy2020image} directly tokenizes and flattens 2D images into a sequence of tokens. Since then, researchers have been working on improving Vision Transformers and examples include DeiT~\cite{touvron2020training}, T2T-ViT~\cite{yuan2021tokens}, PVT~\cite{wang2021pyramid}, and Swin-Transformer~\cite{liu2021swin}. In addition, the intriguing properties of ViT are investigated in~\cite{IPVT}.
	
	\noindent \textbf{Jigsaw puzzle solver.} 
	The goal of a jigsaw puzzle solver is to reconstruct an original image from its shuffled patches. Since this problem is NP-hard~\cite{Demaine}, solving jigsaw puzzles of non-trivial size is impossible. Most of the existing works in computer vision focus on the jigsaw puzzle problem composed of equally-sized image patches and examples include the greedy algorithm proposed in~\cite{5probabilistic_jigsaw}, the particle filter-based algorithm proposed in~\cite{Alpher26}, the fully-automatic solver proposed in ~\cite{17fully_automated}, and the genetic-based solver proposed in~\cite{Alpher29}.

	\noindent \textbf{Privacy-preserving machine learning.} 
	The aim of privacy-preserving machine learning is to integrate privacy-preserving techniques into the machine learning pipeline. According to the phases of privacy integration, existing methods can be basically divided into four categories: data preparation, model training and evaluation, model deployment, and model inference~\cite{MethodsChallengesDirections}. Federated learning allows multiple participants to jointly train a machine learning model while preserving their private data from being exposed~\cite{DropoutResilient}. However, the leakage of gradients~\cite{GradViT} or confidence information ~\cite{ConfidenceInformation} can be utilized to recover original images such as human faces. From the perspective of data, encrypting data and then learning and inferencing on encrypted data can provide a strong privacy guarantee, called confidential-level privacy, which receives increasing attention recently. Manually anonymizing large-scale datasets faces the challenge of inefficiency and the need to develop specialized techniques~\cite{ijcv2}, and thus the automatic encryption approaches such as homomorphic encryption and functional encryption have been employed to encrypt data due to their nature of allowing computation over encrypted data~\cite{CryptoNN, EncryptedData}. However, scaling them to deep networks and large datasets still faces extreme difficulties due to the high computational complexity involved for encryption. The block-wise pixel shuffling encryption strategy~\cite{2018-ICCETW-Tanaka, madono2020block} and the pixel-based image encryption strategy~\cite{Warit_IEEEtrans} have low computational complexity for encryption, but they are vulnerable to reconstruction attacks because of the limited dimension of the key space~\cite{Warit_IEEEtrans} or suffer from large performance degradation~\cite{madono2020block}. In contrast, our encryption strategies are simple, efficient and easy to implement, and can be applied to position-sensitive tasks.
	
	\noindent \textbf{Differential privacy.} Differential privacy is defined in terms of the application-specific concept of adjacent datasets~\cite{DifferentialPrivacy}, which bounds the disclosure risk of any individual participating in a dataset to guarantee data privacy. Then, researchers have been working on expanding its application scope. For example, Ferdinando et al. propose a differential-privacy mechanism for releasing hierarchical counts of individuals~\cite{FIORETTO2021103475}. Liu et al. study the interpretation robustness problem from the perspective of R$\acute{e}$nyi differential privacy. We aim to process images to be ``human-imperceptible, machine-recognizable'', which is different from differential privacy that limits the information that attackers can learn about datasets. Similar idea is also mentioned in~\cite{pmlr-v119-huang20i} but the scheme has proved to be not private in~\cite{NeuraCrypt}.
	
	\noindent \textbf{Word order in transformers for NLP.} Recent studies in the natural language processing field suggest that higher-order co-occurrence statistics of words play a major role in masked language models like BERT~\cite{wordorderlittle}. Moreover, it has been shown that the word order contains surprisingly little information compared to that contained in the bag of words, since the understanding of syntax and the compressed world knowledge held by large models (e.g. BERT and GPT-2) are capable to infer the word order~\cite{bagofwords}. Due to the property of attention operation, when removing positional encoding, ViT is permutation-invariant w.r.t. its attentive tokens. As evaluated by our experiments, removing the positional embedding from ViT only leads to a moderate performance drop (\%3.1, please see Table~\ref{exp:imagenet}). Such a phenomenon inspires us to explore the permutation-based encryption strategy.

	\section{Method}
	In this section, we propose an efficient privacy-preserving learning paradigm that can ensure the encrypted images have become human-imperceptible while preserving machine-recognizable information. We first provide the encryption strategies in Sec.~\ref{subsec:encryption} and then detail the building blocks of ViT in Sec.~\ref{subsec:blocks}. Next, we describe how to learn on the encrypted images in Sec.~\ref{subsec:method-classification} and Sec.~\ref{subsec:method-obeject-deteciton} with minimal modifications to ViT. Finally, we discuss the encryption strategies in the context of cryptography in Sec.~\ref{subsec:cryptography}.
	
	\subsection{Human-imperceptible images}
	\label{subsec:encryption}
	
	We first consider typical vision tasks which are not quite position-sensitive, such as image classification that predicts the global category. Here, \textit{Random Shuffling (RS)} images to a set of equally-sized image patches can destroy human-recognizable contents. This shuffle-encrypted process is simple, easy to implement, and is decoupled from the network optimization. Under the circumstance, decrypting an image is solving a jigsaw puzzle problem, which can incur a large computational overhead since the problem to be solved is an NP-hard one~\cite{Demaine}. In particular, the dimension of the key space when applying the shuffle-encrypted strategy is the number of puzzle permutations. For an image with $N$ patches, the dimension of the key space is given by,
	\begin{equation}
	K_S = N!\ .
	\end{equation}
	For example, a $7\times7$ puzzle has $49! \approx 6 \times 10^{62}$ possible permutations. Based on this, it is easy to further increase the complexity of decrypting an image, by reducing the patch size of puzzles or increasing the resolution of the image. Besides, the complexity of decrypting an image can be further increased by dropping some image patches, as experimentally evaluated in Figure~\ref{fig:attacks} and Figure~\ref{fig:attacks_size}.
	
	However, always shuffling image patches is prone to underestimating the position information. To further make our paradigm applicable to position-sensitive tasks like object detection, where precise positions of bounding boxes are predicted, we design another encryption strategy named  \textit{Mixing (MI)}. Specifically, MI mixes sub-patches of image patches to destroy human-recognizable contents while preserving the position information, so that the encrypted data can be learned by networks that need position information. 
	Assuming that there are $M$ sub-patches extracted from an image patch $x^p$, the mixing-encrypted strategy can be formulated as follows,
	\begin{equation}
	\mbf{x}_S^p = \frac{1}{M} \sum_{i=1}^{M} \mbf{s}_i^p\ ,
	\end{equation}
	where $\mbf{s}_i^p$ denotes the $i$-th sub-patch of $x^p$. Since the $\mathrm{sum}$ function is permutation-invariant, MI makes an encrypted image permutation-invariant to its sub-patches. With the MI encryption process, decrypting a patch is solving the following ill-posed inverse problem,
	\begin{equation}
	\mathop{\arg\min} \limits_{\mbf{s}_1^p, \cdots, \mbf{s}_M^p} \parallel \mbf{x}_S^p - \sum_{i=1}^{M} \mbf{s}_i^p  \parallel^2 \ ,
	\end{equation}
	Both modeling the sub-patch distribution and restoring the sub-patch order make decrypting an patch a great challenge. Decrypting an image of $N$ patches magnifies this challenge by a factor of $N$.
	
	\subsection{Building Blocks of ViT}
	\label{subsec:blocks}
	
	In this part, we analyze how the change of input permutation affects each component of ViT. 
	
	\textbf{Self-attention.} The attention mechanism is a function that outputs the weighted sum of a set of $k$ \emph{value} vectors (packed into $V\in\mathbb{R}^{k\times d}$). The $k$ weights are obtained by calculating the similarity between a \emph{query} vector $q\in\mathbb{R}^d$ and a set of $k$ \emph{key} vectors (packed into $K\in\mathbb{R}^{k\times d}$) using inner products, which are then scaled and normalized with a softmax function. For a sequence of $N$ query vectors (packed into $Q\in\mathbb{R}^{N\times d}$), the output matrix $O$ (of size $N\times d$) can be computed by,
	\begin{equation}
	O = \mathrm{Attention}(Q, K, V) = \mathrm{Softmax}(Q K^\top/\sqrt{d}) V,
	\end{equation}
	where the $\mathrm{Softmax}$ function is applied to the input matrix by rows.
	
	In self-attention, \emph{query}, \emph{key}, and \emph{value} matrices are computed from  the same sequence of $N$ input vectors (packed into $X\in \mathbb{R}^{N\times d}$) using linear transformations: $Q=XW^\mathrm{Q}$, $K=XW^\mathrm{K}$, $V=XW^\mathrm{V}$. Since the order of $Q$, $K$, and $V$ is co-variant with that of $X$,
	the permutation of the input of self-attention permutes the output.
	
	\textbf{Multi-head self-attention.} Multi-head Self-Attention (MSA)~\cite{vaswani2017attention} consists of $h$ self-attention layers, each of which outputs a matrix of size $N\times d$. These $h$ output matrices are then rearranged into a $N \times dh$ matrix that is reprojected by a linear layer into $N \times d$,
	\begin{equation}
	\mathrm{MSA} = \mathrm{Concat}(\mathrm{head}_1, \cdots, \mathrm{head}_h)W^\mathrm{O},
	\end{equation}
	where $\mathrm{head}_i=\mathrm{Attention}(Q_i, K_i, V_i)$, $Q_i=XW^\mathrm{Q}_i$, $K_i=XW^\mathrm{K}_i$, $V_i=XW^\mathrm{V}_i$, and $W^\mathrm{O} \in\mathbb{R}^{dh \times d}$. The order of each head is co-variant with that of $X$. Since the $\mathrm{Concat}$ operation concatenates the vectors from different heads at the same position, it is co-variant to the order of $X$. Therefore, we conclude that the permutation of the input of MSA permutes the output.
	
	\textbf{Layer normalization.} Layer normalization~\cite{LayerN} is applied to the last $d$ dimensions, based on which the mean and standard deviation are calculated. Therefore, the permutation of the input of layer normalization permutes the output.
	
	\textbf{Residual connection.} The residual connection~\cite{he2016deep} can be formulated as $\mathcal{F}(X) + X$. If the order of the nonlinear mapping $\mathcal{F}(\cdot)$ is co-variant with that of $X$, the residual connection is co-variant with that of $X$.
	
	\textbf{Feed-forward network.} Feed-Forward Network (FFN) consists of two linear layers separated by a GELU activation~\cite{gelu}. The first linear layer expands the dimension from $D$ to $4D$, while the second layer reduces the dimension from $4D$ back to $D$. Since FFN is applied on the last $d$ dimensions, the permutation of the input of FFN permutes the output.
	
	\textbf{Positional encoding.}  To retain the positional information, position embeddings are usually added to the patch embeddings. Since there is little to no difference between different ways of encoding positional information, learnable 1D position embeddings are used in ViT~\cite{dosovitskiy2020image}. It is to be noted that the positional encoding is unaware of input permutation.
	
	\subsection{Classification on Encrypted Images}
	\label{subsec:method-classification}
	
	As a standard method to handle images in ViT, the fixed-size input image of $H\times W\times C$ is decomposed into a batch of $N$ patches of a fixed resolution of $P\times P$, resulting in a sequence of flattened 2D patches $X^p \in \mathbb{R}^{N \times (P^2 \cdot C)}$. For example, the sequence length $N=HW/P^2$ of ViT could be 196 for image classification on the ImageNet dataset. To destroy the human-recognizable contents, we choose RS as the encryption strategy to encrypt images. The reason is two-fold: (1) The key space of an image encrypted by RS is big enough and (2) The drop in performance is insignificant. To learn on the images encrypted by RS, we design permutation-invariant ViT (PEViT), defined as follows,
	\begin{align}
	\{\mbf{x}_1^p& , \mbf{x}_2^p, \cdots, \mbf{x}_N^p\} \xrightarrow{\text{Shuffling}} \{\mbf{x}_2^p, \mbf{x}_N^p, \cdots, \mbf{x}_1^p\} \label{eq:shuffle} \\
	\mbf{z}_0 &= [ \mbf{x}_\text{class}; \, \mbf{x}_2^p \mbf{E}; \, \mbf{x}_N^p \mbf{E}; \cdots; \, \mbf{x}_{1}^p \mbf{E} ],
	&& \mbf{E} \in \mathbb{R}^{(P^2 \cdot C) \times D} \label{eq:embedding} \\
	\mbf{z^\prime}_\ell &= \op{MSA}(\op{LN}(\mbf{z}_{\ell-1})) + \mbf{z}_{\ell-1}, && \ell=1\ldots L \label{eq:msa_apply} \\
	\mbf{z}_\ell &= \op{FFN}(\op{LN}(\mbf{z^\prime}_{\ell})) + \mbf{z^\prime}_{\ell}, && \ell=1\ldots L  \label{eq:mlp_apply} \\
	\mbf{y} &= \op{LN}(\mbf{z}_L^0). \label{eq:final_rep}
	\end{align}
	where $\mbf{E}$ denotes the linear projection that maps each vectorized image patch to the model dimension $D$, and $\mbf{x}_\text{class}$ denotes the class token ($\mbf{z}_0^0=\mbf{x}_\text{class}$), whose state at the output of the visual transformer ($\mbf{z}_L^0$) serves as the image representation $\mbf{y}$.
	
	The differences between PEViT and vanilla ViT are two-fold: (1) Our model takes shuffled patch embeddings as input and (2) The learned positional encodings are removed. Since the permutation of the input of all the building blocks permutes the output, the order of $\mbf{z}_L$ is co-variant with that of $\mbf{z}_0$. It is worth noting that the class token is fixed in $\mbf{z}_0^0$. Therefore, $\mbf{z}_L^0$ corresponds to the image representation $\mbf{y}$ that is invariant to the order of patch embeddings or image patches.
	
	\begin{figure*}[t]
		\centering
		\includegraphics[width=0.95\linewidth]{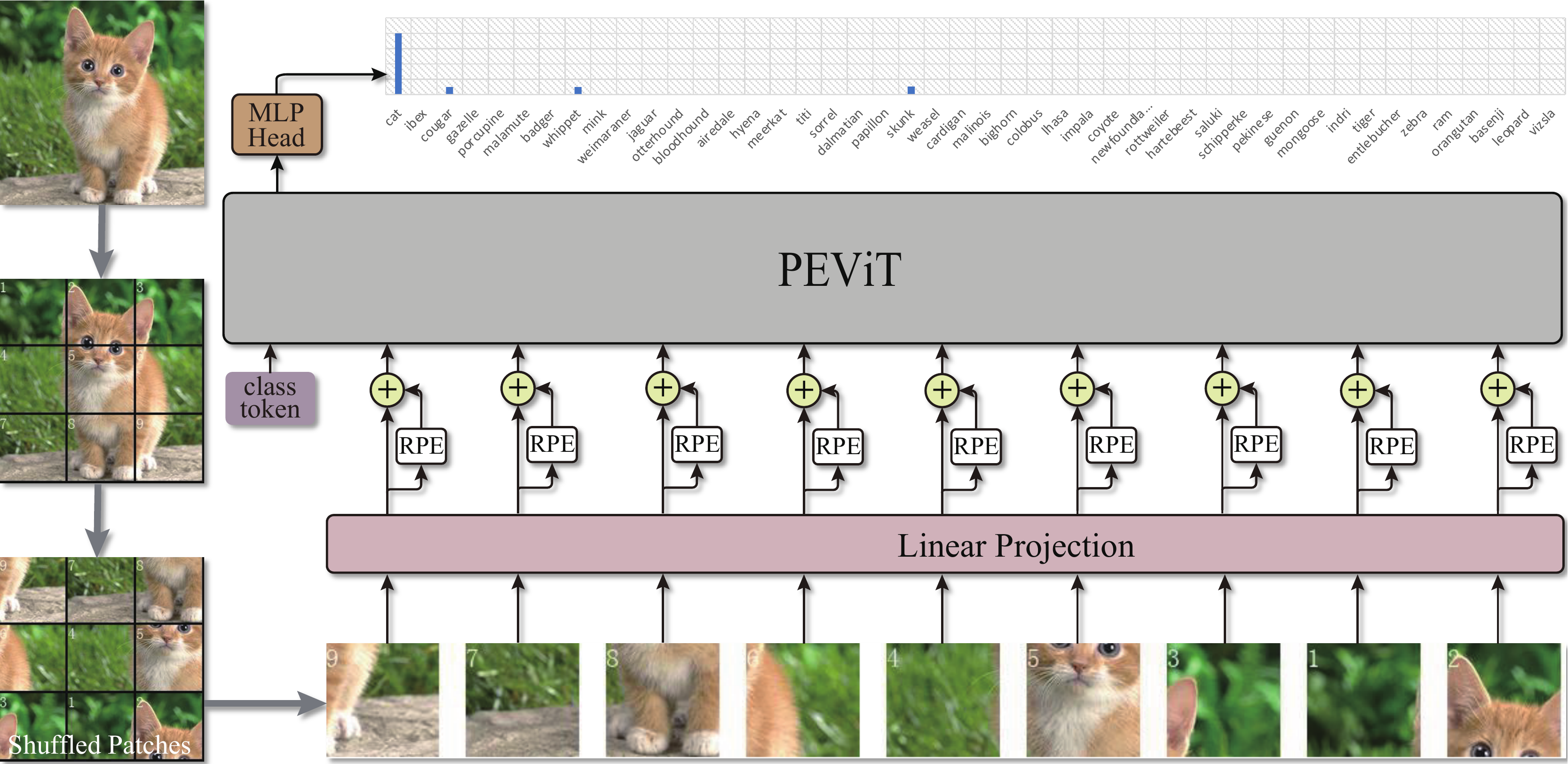}
		\vspace{-0.0em}
		\caption{Architecture overview of PEViT with RPE.}
		\label{fig:pipline_cls}
	\end{figure*}
	
	When directly introducing positional encoding, the permutation-invariant property of PEViT is destroyed. Inspired by relative encoding~\cite{RelativePosition}, we propose a reference-based positional embedding approach that can retain the permutation-invariant property,
	\begin{equation}
	\mbf{E}_{pos}(\mbf{x}_i^p) = \mathrm{RPE}(\mbf{x}_i^p - \mbf{x}^{\text{ref}}),
	\end{equation}
	where $\mbf{x}^{\text{ref}} \in \mathbb{R}^{d}$ denotes the learnable reference embedding and $\mathrm{RPE}$ denotes the reference-based positional encoding network that consists of two linear layers separated by a GELU activation~\cite{gelu}, followed by a sigmoid function. This reference-based positional embedding relies only on the learnable reference embedding and patch embeddings, and thus its order is co-variant with that of input vectors. Accordingly, the input to PEViT with RPE can be defined as follows,
	\begin{equation}
	\mbf{z}_0 = [ \mbf{x}_\text{class}; \, \mbf{x}_2^p \mbf{E} + \mbf{E}_{pos}(\mbf{x}_2^p); \, \mbf{x}_N^p \mbf{E} + \mbf{E}_{pos}(\mbf{x}_N^p); \cdots; \, \mbf{x}_{1}^p \mbf{E} + \mbf{E}_{pos}(\mbf{x}_1^p) ],
	\end{equation}
	where the permutation-invariant property of PEViT is retained. An architecture overview of PEViT with RPE is depicted in Figure~\ref{fig:pipline_cls}.
	
	Through the reference-based positional encoding, we illustrate that introducing positional embedding while retaining permutation-invariant property is feasible. Other better positional embedding approaches can also be designed, but it is beyond the scope of this paper.
	
	\begin{figure*}[t]
		\centering
		\includegraphics[width=0.95\linewidth]{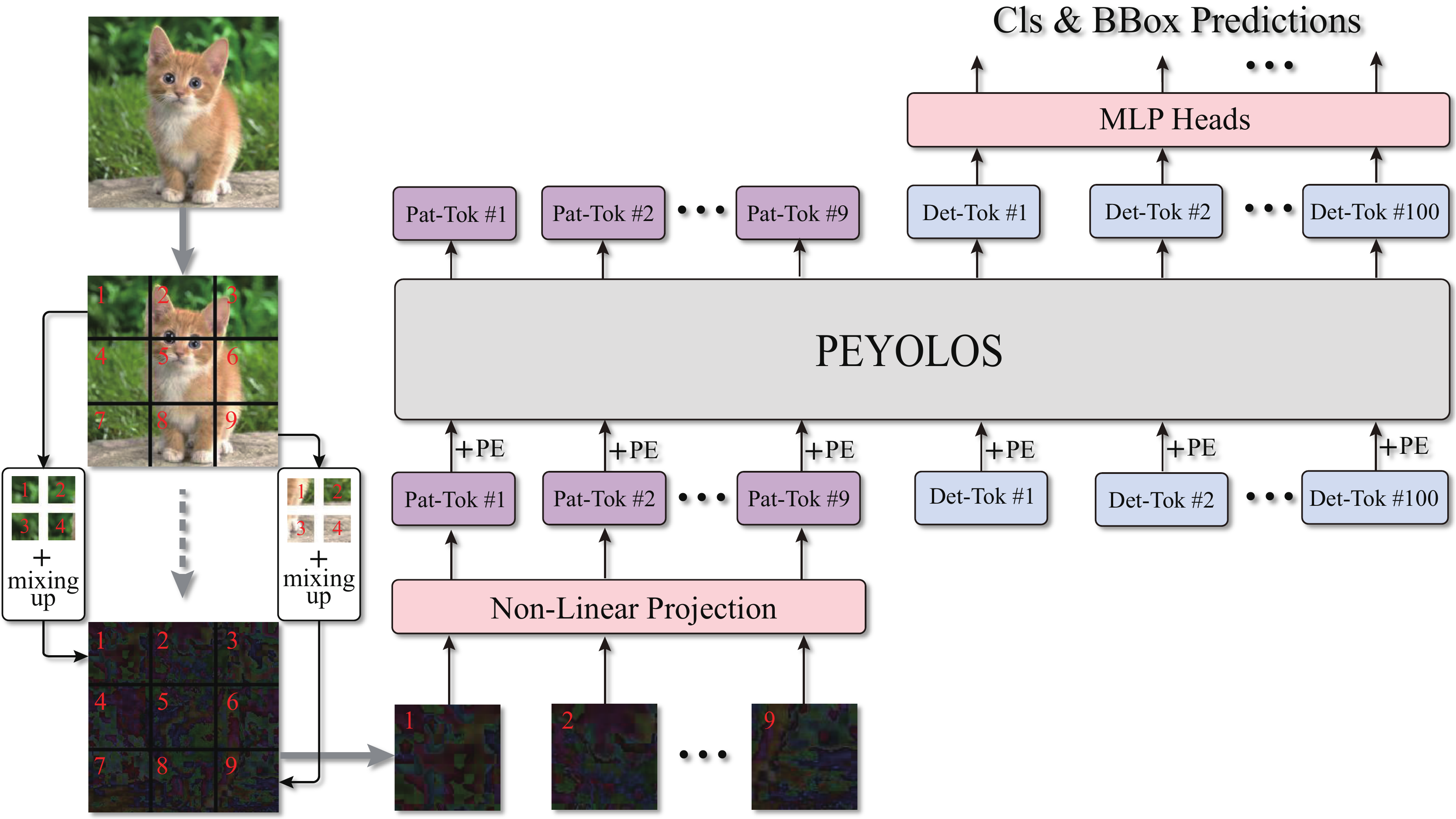}
		\vspace{-0.0em}
		\caption{Architecture overview of PEYOLOS.}
		\label{fig:det}
	\end{figure*}
	
	\subsection{Object Detection on Encrypted Images}
	\label{subsec:method-obeject-deteciton}
	
	YOLOS~\cite{YOLOS} is an object detection model based on the vanilla ViT. The change from a ViT to a YOLOS involves two steps: (1) Dropping the class token and appending 100 randomly initialized learnable detection tokens to the input patch embeddings and (2) Replacing the image classification loss with the bipartite matching loss to perform object detection in a set prediction manner~\cite{carion2020end}.
	
	Since position information plays a key role in the low-level object detection task, directly encrypting images with RS disrupts the patch positions and thus leads to significant performance degradation. This brings a great challenge to destroy human-recognizable contents while preserving machine-learnable information for the object detection task. We address this challenge by adapting the way image patches are embedded. For an image patch ($\mbf{x}_i^p$) of a fixed resolution $P \times P$, we further decompose the patch into a batch of 4 sub-patches of a fixed resolution $\frac{P}{2} \times \frac{P}{2}$. Then, these sub-patches are encrypted with MI, resulting in the encrypted patch $\mbf{x}_i^S$. Accordingly, the input to YOLOS is adapted as follows,
	\begin{equation}
	\mbf{z}_0 = [ \mbf{x}_\text{1}^{DET}; \cdots; \mbf{x}_\text{100}^{DET}; \, \mathcal{H}(\mbf{x}_1^S); \mathcal{H}(\mbf{x}_2^S); \cdots; \mathcal{H}(\mbf{x}_N^S)] + \mbf{P},
	\end{equation}
	where $\mathcal{H}$ denotes a nonlinear mapping composed of two linear layers separated by a GELU activation~\cite{gelu} and $\mbf{P} \in \mathbb{R}^{(100 + N) \times D}$ denotes the learnable positional embeddings. This adaptation makes YOLOS partially permutation-invariant to its sub-patches. With our PEYOLOS, we can destroy human-recognizable contents while preserving machine-learnable information for the object detection task. 
	
	The pipeline of PEYOLOS is shown in Figure~\ref{fig:det}. The input image of $H\times W\times C$ is decomposed into a batch of $N$ patches with a fixed resolution of $P\times P$. Then, for an image patch, we further decompose the patch into a batch of 4 sub-patches with a fixed resolution $\frac{P}{2} \times \frac{P}{2}$. Finally, these sub-patches are encrypted with MI, resulting in a highly encryted image that is not human-recognizable and is tough to be decrypted. PEYOLOS takes as input the encrypted images and outputs class and bounding box predictions for the original images.
	
	\subsection{Discussion}
	\label{subsec:cryptography}
	
	We would like to recall a framework termed substitution-permutation network (SPN) in cryptography~\cite{Cryptography}. SPN is a series of linked mathematical operations used in block cipher algorithms such as AES and DES~\cite{Cryptography}. Shannon suggests that practical and secure ciphers may be constructed by employing a mixing transformation consisting of several rounds of confusion and diffusion~\cite{6769090}; SPN is exactly an implementation of this confusion and diffusion paradigm. Such an implementation applies several alternating rounds of substitution boxes (S-boxes) and permutation boxes (P-boxes) to produce the ciphertext. An S-box substitutes a small block of bits (the input of the S-box) by another block of bits (the output of the S-box). A P-box is a permutation of bit blocks (or bits). Although a single typical S-box or a single P-box alone does not have sufficient cryptographic strength, a well-designed SPN with several alternating rounds of S-boxes and P-boxes already has a very strong proven security~\cite{Provablesecurity}.
	
	Our encryption scheme adheres to SPN, while the basic unit of our encryption scheme is pixels instead of bits. In particular, MI can be seen as an implementation of S-box. RS can be seen as an implementation of P-box. Although both MI and RS alone does not have sufficient cryptographic strength, alternating several rounds of MI and RS can enhance the cryptographic strength to a large extent; see Figure~\ref{fig:encrypted_images}. We would like to stress out that the performance on encrypted images is also a major concern of our work. Although alternating several rounds of MI and RS can enhance cryptographic strength, it also decreases the performance by a significant margin. In practice, there is a tradeoff between performance and cryptographic strength.
	
	\section{Experiments}
	
	In this section, we first provide the experimental settings, and then contrast the performance of the proposed PEViT and PEYOLOS on image classification and object detection tasks. Next, we investigate various attackers, including puzzle solver attacker, gradient leakage attacker, and reconstruct attacker. Finally, we ablate the key components. The code is available at \url{https://github.com/FushengHao/PrivacyPreservingML.}
	
	\begin{algorithm}[t]
		\caption{\textcolor{rebuttle}{Pseudocode of RS in a PyTorch-like style.}}
		\label{alg:rs}
		\definecolor{codeblue}{rgb}{0.25,0.5,0.5}
		\lstset{
			backgroundcolor=\color{white},
			basicstyle=\fontsize{7.2pt}{7.2pt}\ttfamily\selectfont,
			columns=fullflexible,
			breaklines=true,
			captionpos=b,
			commentstyle=\fontsize{7.2pt}{7.2pt}\color{codeblue},
			keywordstyle=\fontsize{7.2pt}{7.2pt},
		}
		\begin{lstlisting}[language=python]
		class RS_PatchEmbed(nn.Module):
		# Key implementation details of RS
		def __init__(self, patch_size, in_dim, embed_dim):
		super().__init__()
		patch_size = to_2tuple(patch_size)
		self.proj = nn.Conv2d(in_dim,embed_dim,kernel_size=patch_size,stride=patch_size)
		
		def forward(self, x):
		B, C, H, W = x.shape
		x = self.proj(x).flatten(2).transpose(1, 2)  # B x HW x C
		
		shuffle = []
		for idx in range(B):
		random_idx = torch.randperm(x.size(1))
		shuffle.append(x[idx][random_idx, :].unsqueeze(0))
		
		x = torch.cat(shuffle, dim=0)
		
		return x
		\end{lstlisting}
	\end{algorithm}

	\begin{algorithm}[t]
		\caption{\textcolor{rebuttle}{Pseudocode of MI in a PyTorch-like style.}}
		\label{alg:mi}
		\definecolor{codeblue}{rgb}{0.25,0.5,0.5}
		\lstset{
			backgroundcolor=\color{white},
			basicstyle=\fontsize{7.2pt}{7.2pt}\ttfamily\selectfont,
			columns=fullflexible,
			breaklines=true,
			captionpos=b,
			commentstyle=\fontsize{7.2pt}{7.2pt}\color{codeblue},
			keywordstyle=\fontsize{7.2pt}{7.2pt},
		}
		\begin{lstlisting}[language=python]
		class MI_PatchEmbed(nn.Module):
		# Key implementation details of MI
		# MI is implemented based on the fact that 0.25*W\sum_{i=1}^{4}{x_i}=0.25*\sum_{i=1}^{4}{Wx_i}
		def __init__(self, patch_size, in_dim, embed_dim):
		super().__init__()
		patch_size = to_2tuple(patch_size)
		self.proj_1 = nn.Conv2d(in_dim,embed_dim,kernel_size=patch_size,stride=patch_size)
		self.pool = nn.AvgPool2d(2, stride=2)
		self.act = nn.GELU()
		self.proj_2 = nn.Conv2d(embed_dim, embed_dim, kernel_size=1, stride=1)
		
		def forward(self, x):
		B, C, H, W = x.shape
		x = self.proj_1(x)  # B x C x (H/8) x (W/8)
		x = self.pool(x)  # B x C x (H/16) x (W/16)
		x = self.proj_2(self.act(x)).flatten(2).transpose(1, 2)  # B x HW x C
		
		return x
		\end{lstlisting}
	\end{algorithm}

	\subsection{Experimental settings}
	
	\textbf{Datasets.} For the image classification task, we benchmark the proposed PEViT on ImageNet-1K~\cite{deng2009imagenet}, which contains $\sim$1.28M training images and 50K validation images. For the object detection task, we benchmark the proposed PEYOLOS on COCO~\cite{lin2014microsoft}, which contains 118K training, 5K validation and 20K test images.
	
	\textbf{Implementation details.} The pseudocode of RS and MI in a PyTorch-like style are shown in Algorithm~\ref{alg:rs} and Algorithm~\ref{alg:mi}. We implement the proposed PEViT based on the Timm library~\cite{timm}. We adopt the default hyper-parameters of the DeiT training scheme~\cite{touvron2020training} except setting the batch size to 192 per GPU, where 8 NVIDIA A100 GPUs are used for training.  It is worth noting PEViT (w.o RPE) is equivalent to removing positional embeddings from DeiT. We implement the proposed PEYOLOS based on the publicly released code in~\cite{YOLOS} and adapt the way image patches are embedded.
	
	\textbf{Baseline.} We propose an efficient privacy-preserving learning paradigm with the aim of destroying human-recognizable contents while preserving machine-learnable information. The proposed PEViT and PEYOLOS are inherited from DeiT~\cite{touvron2020training} and YOLOS~\cite{YOLOS} respectively, which are thus selected as baselines. It is worth noting that both PEViT and PEYOLOS are not designed to be high-performance models that beats state-of-the-art image classification and object detection models, but to unveil that destroying human-recognizable contents while preserving machine-learnable information is feasible.
	
	\textbf{Measurement of privacy protection.} As shown in Figure~\ref{fig:encrypted_images}, the visual contents of encrypted images are nearly-completely protected from recognizing by human eyes. To measure the strength of privacy protection, we try to restore the original images with various attack algorithms, including puzzle solver attacker, gradient leakage attacker, and reconstruct attacker. Then, the quality of restored images can reflect the strength of privacy protection.
	
	\begin{table*}[!t]
		\setlength\tabcolsep{3.5pt}
		\centering
		\caption{Comparison of different methods on ImageNet-1K classification.}
		\label{exp:imagenet}
		\small
		\begin{tabular}{l|ccc|c}
			method &  \begin{tabular}[c]{@{}c@{}}image \\ size\end{tabular} & \#param. & FLOPs & \begin{tabular}[c]{@{}c@{}}ImageNet \\ top-1 acc.\end{tabular} \\
			\hline
			RegNetY-16G~\cite{radosavovic2020designing} & 224$^2$ & 84M & 16.0G & 82.9 \\
			EffNet-B3~\cite{tan2019efficientnet} & 300$^2$ & 12M & 1.8G & 81.6 \\
			EffNet-B4~\cite{tan2019efficientnet} & 380$^2$ & 19M & 4.2G & 82.9 \\
			EffNet-B5~\cite{tan2019efficientnet} & 456$^2$ & 30M & 9.9G & 83.6 \\
			DeiT-B~\cite{touvron2020training} & 224$^2$ & 86M & 17.5G &  81.8 \\
			TNS-B~\cite{han2021transformer} & 224$^2$ & 66M & 14.1G &  82.8 \\
			T2T-ViT-14~\cite{yuan2021tokens} & 224$^2$ & 22M & 5.2G &  81.5 \\
			T2T-ViT-24~\cite{yuan2021tokens} & 224$^2$ & 64M & 14.1G &  82.3 \\
			PVT-Large~\cite{wang2021pyramid} & 224$^2$ & 61M & 9.8G & 81.7 \\
			Swin-T~\cite{liu2021swin} & 224$^2$ & 28M & 4.5G & 81.2 \\
			Swin-S~\cite{liu2021swin} & 224$^2$ & 50M & 8.7G & 83.2 \\
			GG-T~\cite{Glance-and-Gaze} & 224$^2$ & 28M & 4.5G & 82.0 \\
			GG-S~\cite{Glance-and-Gaze} & 224$^2$ & 50M & 8.7G & 83.4 \\
			ViT-B/16~\cite{dosovitskiy2020image} & 384$^2$ & 86M & 55.4G & 77.9 \\
			ViT-L/16~\cite{dosovitskiy2020image} & 384$^2$ & 307M & 190.7G & 76.5 \\
			\hline
			\textbf{DeiT-B} on images encrypted by MI & 224$^2$ & 86M & 17.5G & 78.0 \\
			\hline
			\textbf{PEViT-B} on images encrypted by RS & 224$^2$ & 86M & 17.5G & 78.7 \\
			\textbf{PEViT-B} on images encrypted by RS + MI & 224$^2$ & 86M & 17.5G & 69.5 \\
			\textbf{PEViT-B} with RPE on images encrypted by RS & 224$^2$ & 87M & 17.6G & 79.7 \\
		\end{tabular}
	\end{table*}
	
	\begin{table*}[!t]
		\setlength\tabcolsep{2.5pt}
		\centering
		\caption{Object detection performance on the COCO test2017 dataset. FPS is measured with batch size 1 on a single 1080Ti GPU.}
		\label{tab:detection}
		\small
		\begin{tabular}{l|l|l|c|c|c|c}
			method & backbone  & size & AP & params. (M) & FLOPs (G) & FPS \\ \hline
			YOLOS-Ti & DeiT-Ti & 512 $\times \ *$ & 28.7  & 6.5  & 18.8 & 60  \\
			Deformable DETR   & FBNet-V3  & 800 $ \times \ *$  & 27.9 & 12.2 & 12.3 & 35 \\
			YOLOSv4-Tiny & COSA & 416 $\times$ 416  & 21.7 & 6.1 & 7.0 & 371 \\ 
			CenterNet & ResNet-18 & 512 $\times$ 512  & 28.1 & - & - & 129 \\
			\textbf{PEYOLOS} &  DeiT-Ti  & 512 $\times\ *$     & 25.3 & 7.1  & 19.0 & 58 \\ \hline
			DETR & ResNet-18-DC5 & 800 $\times \ *$ & 36.9  & 29  & 129 & 7.4  \\
			YOLOS-S & DeiT-S & 800 $\times \ *$ & 36.1  & 31  & 194 & 5.7  \\
			\textbf{PEYOLOS} & DeiT-S & 800 $\times \ *$ & 32.9  & 31.6  & 194.9 & 5.6  \\ \hline
			DETR & ResNet-101-DC5 & 800 $\times \ *$ & 42.5  & 60  & 253 & 5.3  \\
			YOLOS-B & DeiT-B & 800 $\times \ *$ & 42.0  & 127  & 538 & 2.7  \\
			\textbf{PEYOLOS} & DeiT-B & 800 $\times \ *$ & 39.5  & 128.2  & 539.7 & 2.5  \\
		\end{tabular}
	\end{table*}
	
	\subsection{Comparison results}
	
	\textbf{ImageNet-1K Classification.}
	The summary of the image classification results is shown in Table~\ref{exp:imagenet}. It can be observed that (1) Our performance drop is only 3.9\% compared with the state-of-the-art CNN-based method EffNet-B5, (2) Our performance drop is only 2.1\% compared with the baseline method DeiT-B, and (3) MI is applicable to image classification. It is worth noting that although the performance of PEViT-B is not state-of-the-art, it is applied on encrypted images where the visual contents of images cannot be recognized by human eyes, while the comparison methods cannot. Moreover, RS and MI can be combined to further improve data security but at the expense of performance. Therefore, we conclude that PEViT-B achieves a trade-off between performance and visual content protection.
	
	\textbf{COCO Object Detection.}
	The summary of the object detection results is shown in Table~\ref{tab:detection}, in which we compare PEYOLOS with the competitive methods that contains roughly equal parameters. It can be observed that PEYOLOS is comparable to these methods. In particular, our performance drop is only $\sim$3.0\% compared with the baseline method YOLOS, and only 2.8\% compared with the CNN-based method CenterNet. Similar to PEViT, although PEYOLOS does not outperform the state-of-the-art methods, it is currently the unique model to achieve a trade-off between performance and visual content protection for object detection. It is worth noting that MI is not limited to YOLOS. With minimal adaptations, other object detection frameworks based on plain Vision Transformer (ViT) can also be adapted to work on images encrypted by MI. Recently, it has been shown that, with ViT backbones pre-trained as Masked Autoencoders, ViTDet~\cite{PlainVisionTransformer} can compete with the previous leading methods that were all based on hierarchical backbones. By adapting the way the image patches are  mapped like PEYOLOS, PEViTDet can be readily obtained, achieving an AP of 41.2.
	
	
	\subsection{Attackers}
	\textbf{Puzzle Solver Attacker.}
	Jigsaw puzzle solver, i.e., reconstructing the image from the set of shuffled patches, can be used to attack the images encrypted with MS. Since this problem is NP-hard~\cite{Demaine}, solving jigsaw puzzles of non-trivial size is impossible. Most of the existing works in computer vision focus on the jigsaw puzzle problem composed of equally-sized image patches~\cite{5probabilistic_jigsaw, 17fully_automated, Alpher29, puzzlessolver}, in which only pixels that are no more than two pixels away from the boundary of a piece are utilized. Therefore, with adaptations on how the image is decomposed, these methods might fail.
	
	\begin{figure*}[!t]
		\centering
		\includegraphics[width=0.95\linewidth]{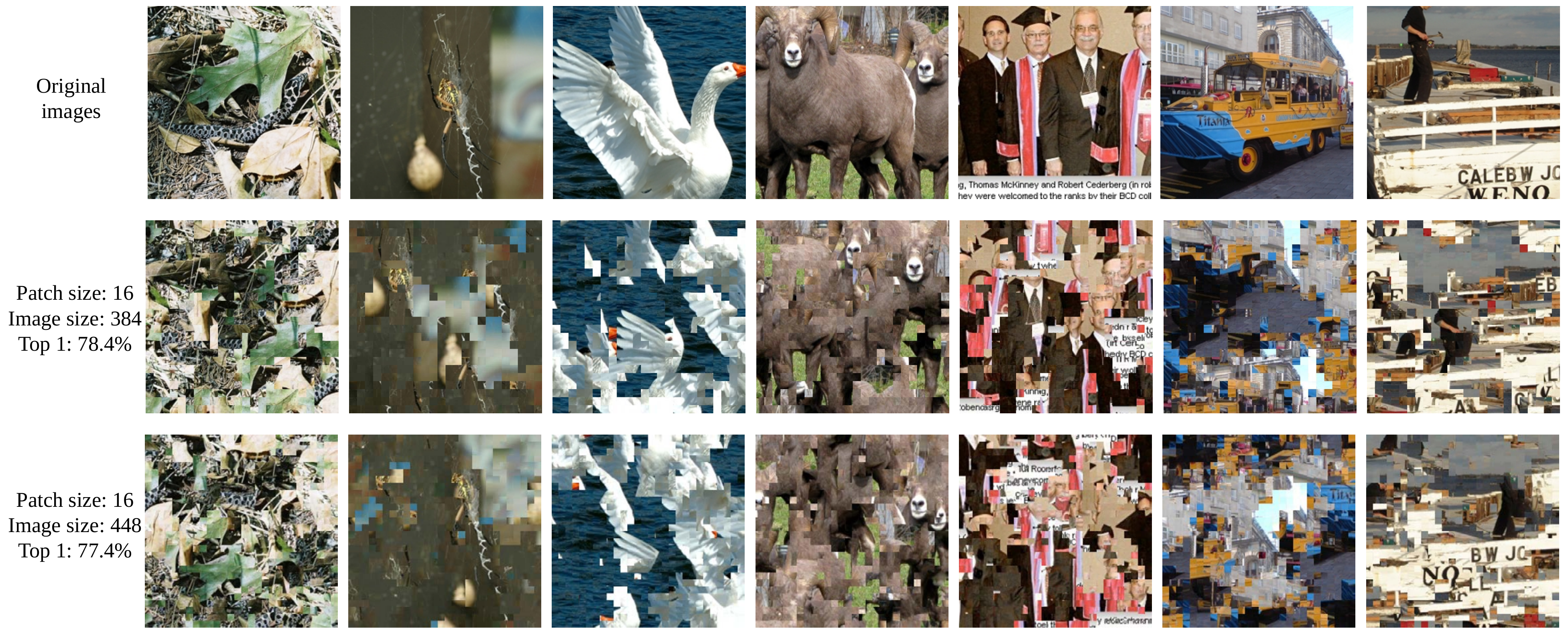}
		\vspace{-0.0em}
		\caption{Reconstructed images by the jigsaw puzzle solver proposed in~\cite{puzzlessolver}. Here, the effect of image size on image reconstruction quality and classification performance is investigated.}
		\label{fig:attacks_size}
	\end{figure*}
	
	\begin{figure*}[!h]
		\centering
		\includegraphics[width=0.95\linewidth]{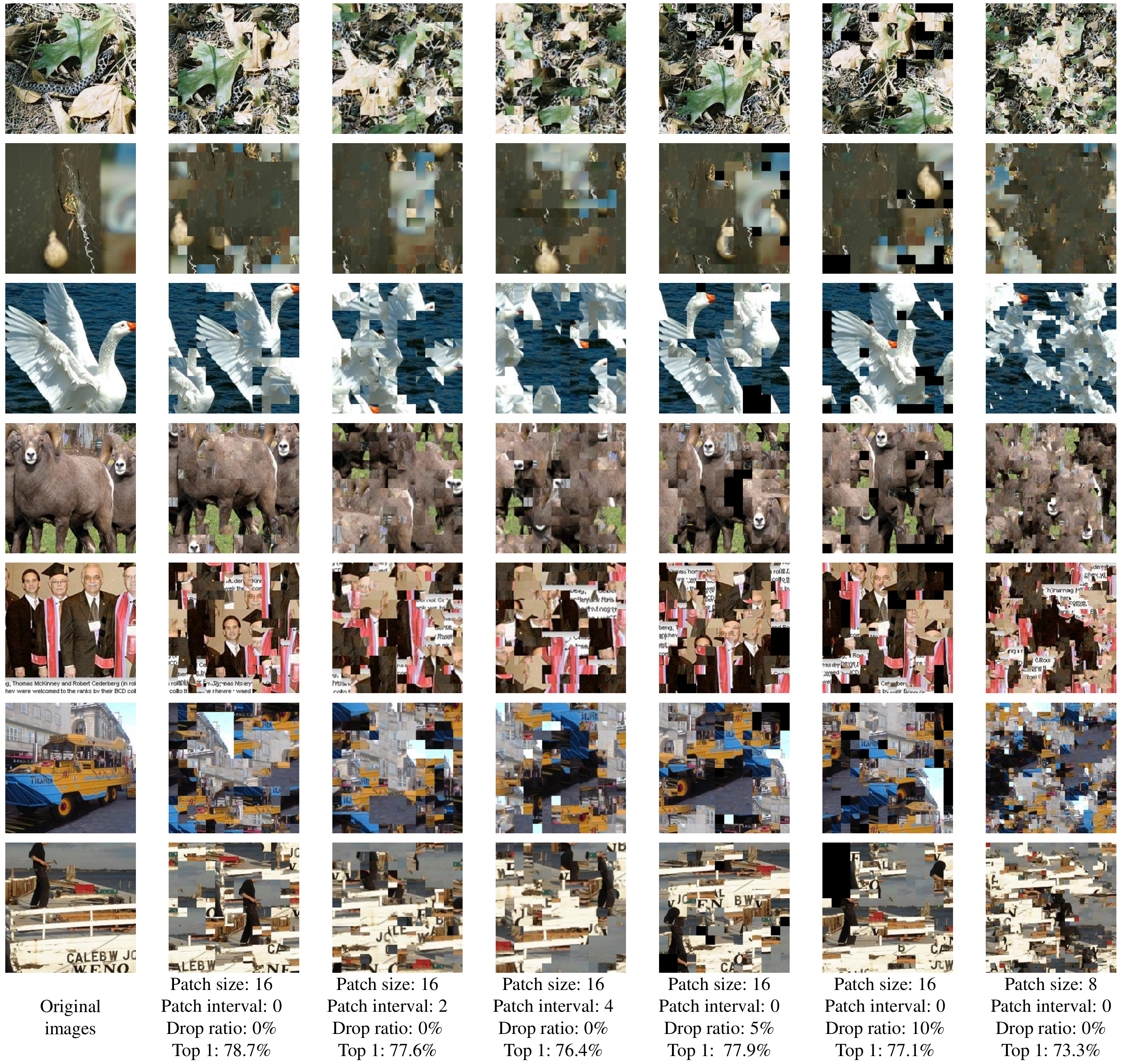}
		\vspace{-0.0em}
		\caption{Reconstructed images by the jigsaw puzzle solver proposed in~\cite{puzzlessolver}, where the default image size is $224\times 224$. Here, the effect of patch size, patch interval, and patch drop ratio on image reconstruction quality and classification performance is investigated.}
		\label{fig:attacks}
	\end{figure*}
	
	We choose the solver proposed in~\cite{puzzlessolver} to inveterate the effect of attacks. The reason is two-fold: (1) It is a fast, fully-automatic, and general solver, which assumes no prior knowledge about the original image and (2) It can handle puzzles with missing pieces. We investigate the effect of patch size, patch interval, patch drop ratio, and image size on image reconstruction quality and classification performance. The results are shown in Figure~\ref{fig:attacks} and Figure~\ref{fig:attacks_size}. We observe that increasing the number of patches (reducing the patch size or increasing the image size) or the patch interval can enhance the strength of privacy protection at the cost of performance. More importantly, even if 10\% of the image patches are dropped, the performance is only reduced by 1.6\%, which allows users to drop patches containing sensitive information.
	
	\begin{figure}[!t]
		\centering
		\includegraphics[width=1.0\linewidth]{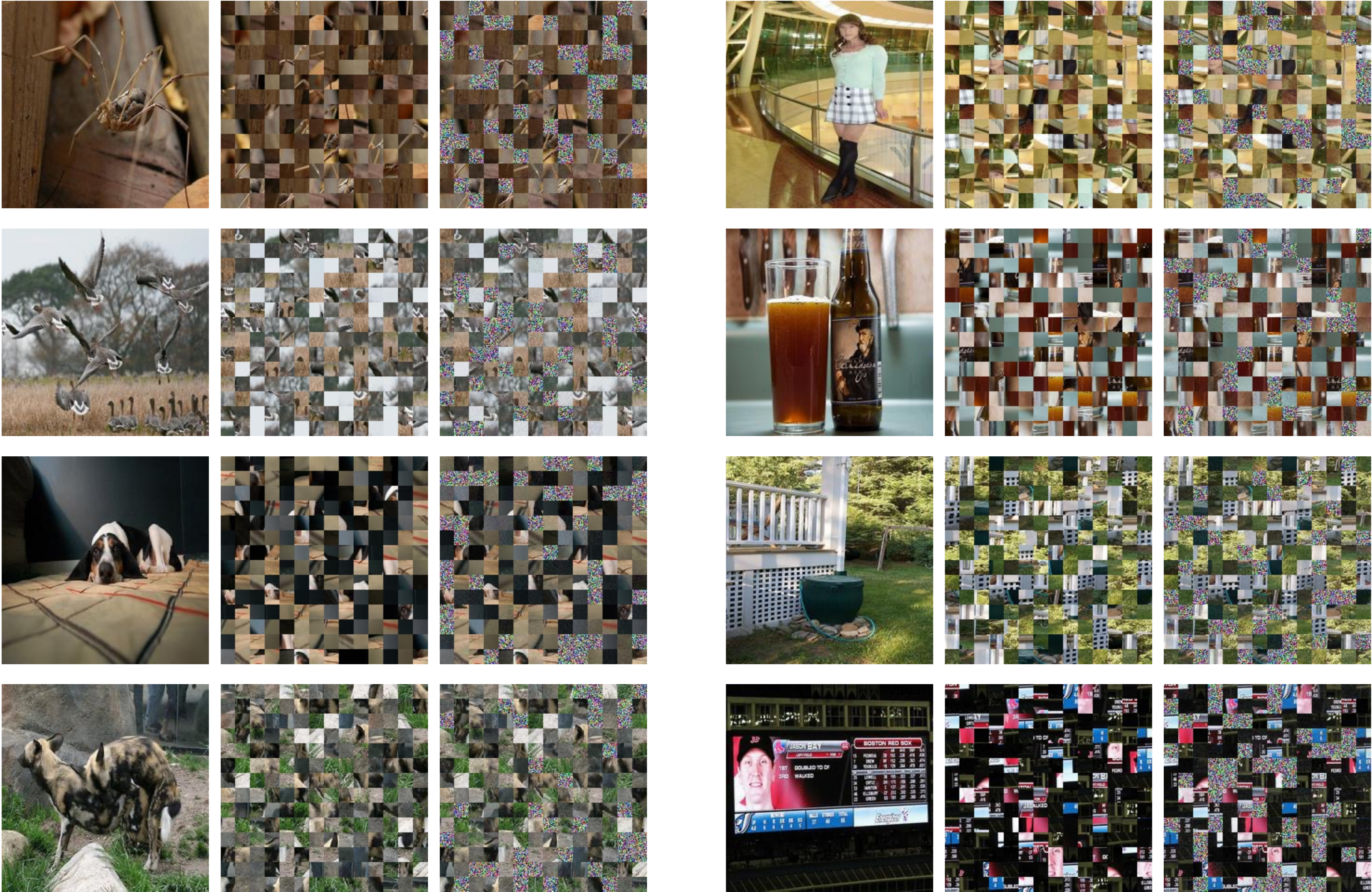}
		\vspace{-1.0em}
		\caption{\textcolor{rebuttle}{Impact of gradient leakage on the security of our method. Left: original images. Middle: images encrypted by RS. Right: recovered images.}}
		\label{fig:grad_leakage}
	\end{figure}
	
	\textbf{Gradient Leakage Attacker.}
	It has been shown that the training set will be leaked by gradient sharing~\cite{DeepLeakage}. To evaluate the impact of gradient leakage on the security of our method, we use the gradient leakage attacker to recover the images encrypted by RS. The results are shown in Figure~\ref{fig:grad_leakage}. It can be observed that (1) Gradient leakage attacker can restore images and (2) the restored images are encrypted. Therefore, we conclude that our paradigm does not prevent gradient leakage attacks, but can make the attacked images useless, thus protecting privacy.
	
\begin{figure}[!t]
		\centering
		\includegraphics[width=0.95\linewidth]{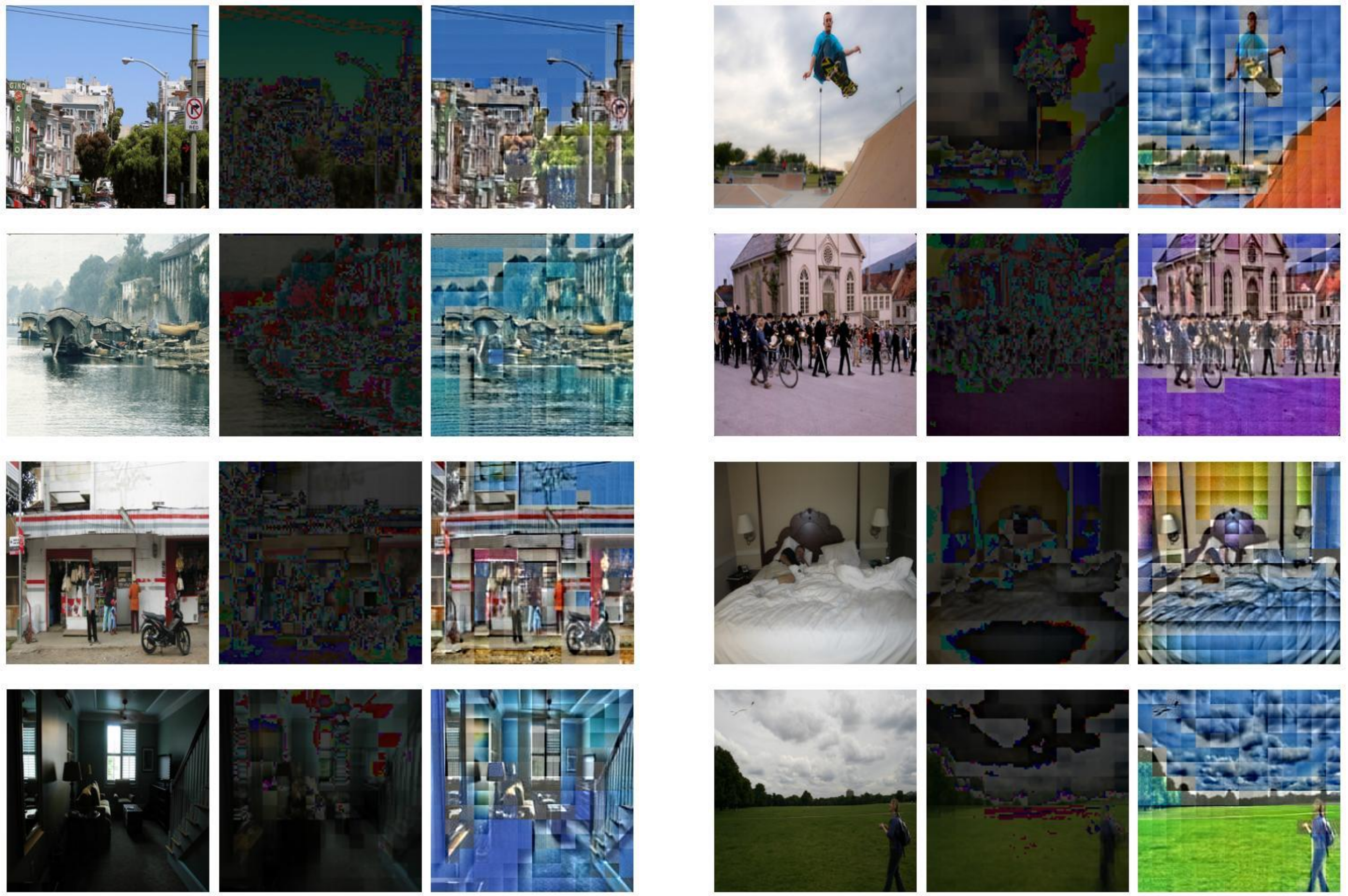}
		\vspace{-0.0em}
		\caption{\textcolor{rebuttle}{Reconstruction attack on MI. Left: original images. Middle: images encrypted by MI. Right: reconstructed images.}}
		\label{fig:mi_attack}
		\vspace{-0.0em}
	\end{figure}

	\textbf{Reconstruction Attacker.}
	\textcolor{rebuttle}{We use a recently proposed powerful Transformer-based framework, MAE~\cite{MAE} (tiny), to recover the original clean images from the images encrypted by MI.}
	\textcolor{rebuttle}{We adapt MAE with two modifications: (1) Patches are not dropped and (2) The linear patch embedding is replaced by a nonlinear patch embedding, which is consistent with the patch embedding used in PEYOLOS. Here the nonlinear patch embedding is composed of two linear layers separated by a GELU activation. Corresponding results are shown in Figure~\ref{fig:mi_attack}. We observe that: (1) The style of reconstructed images is very different from original images and (2) Privacy-sensitive patches such as faces and texts are blurred, and thus the reconstruction with MAE still cannot  reveal the original identity of faces or the contents of texts. These observations indicate that recovering the original clean natural images from images encrypted by MI is a great challenge, demonstrating the effectiveness of MI regarding privacy preserving.}
	
	\begin{figure}[!t]
		\centering
		\includegraphics[width=0.95\linewidth]{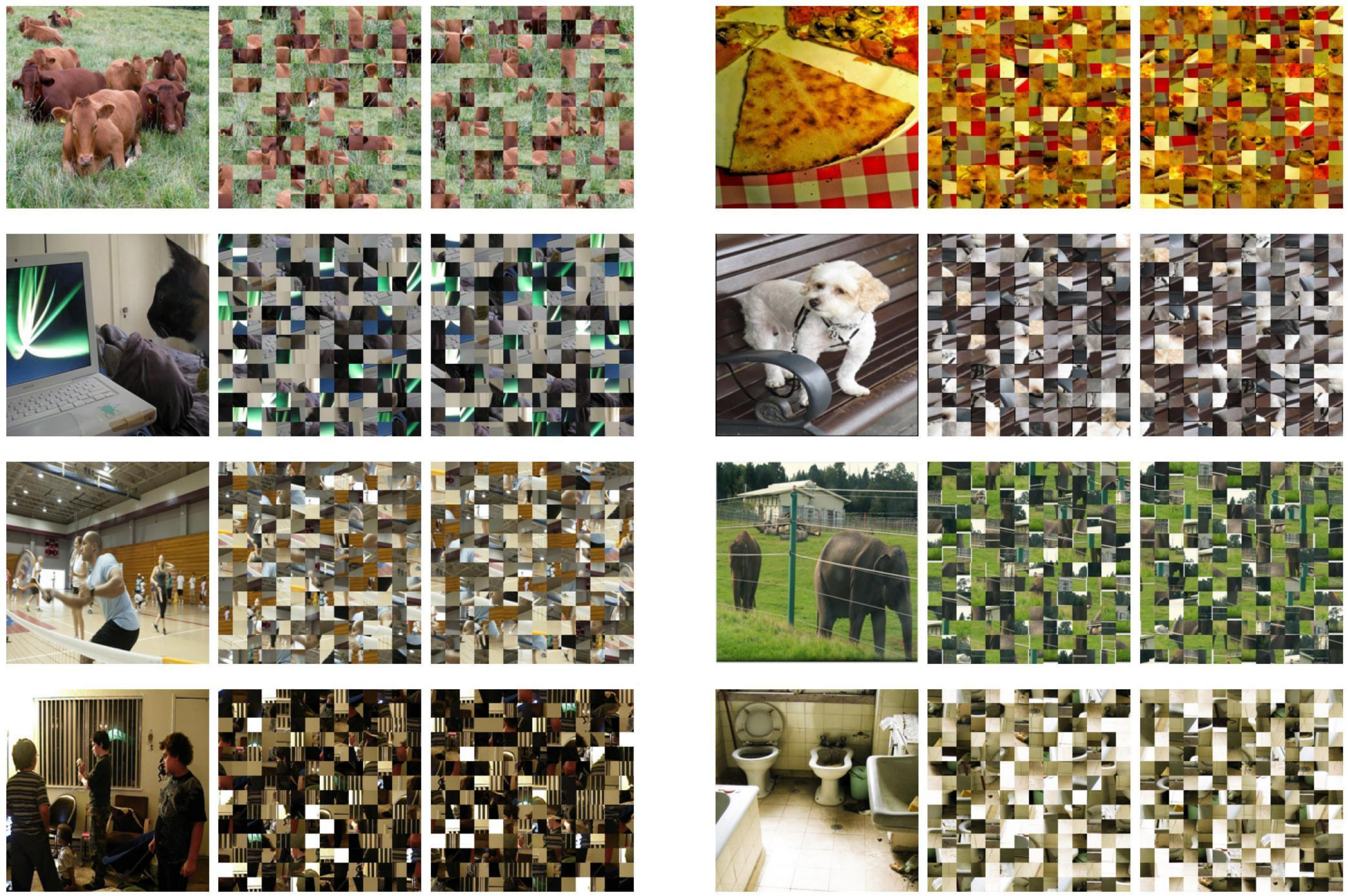}
		\vspace{-0.0em}
		\caption{\textcolor{rebuttle}{Reconstruction attack on RS. Left: original images. Middle: images encrypted by RS. Right: reconstructed images.}}
		\label{fig:rs_attack}
	\end{figure}
	
	\textbf{Transformer-based Puzzle Solver Attacker.}
	\textcolor{rebuttle}{For the large-scale dataset ImageNet, the method in~\cite{Gumbel-Sinkhorn} solves the jigsaw puzzle problem with $3\times 3$ patches. Even in this simple case, the authors still face big challenges. For example, the authors claimed that ``Learning in the Imagenet dataset is much more challenging, as there isn't a sequential structure that generalizes among images''.}
	
	\textcolor{rebuttle}{We extend~\cite{Gumbel-Sinkhorn} to the case where there are $14\times14$ patches (image size $224\times224$, patch size $16\times16$). We use the powerful pipeline of MAE~\cite{MAE} with the following modifications: (1) Patches are not dropped, (2) The loss function used is Gumbel-Softmax proposed in~\cite{Gumbel-Sinkhorn}, and (3) The positional encoding is removed, which is necessary as the patch orders are randomly shuffled by RS. The results are shown in Figure~\ref{fig:rs_attack}.}
	
	
	\textcolor{rebuttle}{It can be observed that predicting the correct patch orders faces big challenges. The reason for this might be that (1) only a tiny fraction of $196! \approx 5\times10^{365}$ possible patch orders are sampled, which is difficult to estimate the underlying joint distribution of patch orders and patch content. Note that sampling a large fraction of $196!\approx5\times10^{365}$ possible patch orders is infeasible in practice. (2) Extremely strong discriminative power are needed. For examples, if we mirror an image, the patch orders should change accordingly. In this case, the input patches are very similar, and a model needs a very strong discriminative power to distinguish them.}
	
	\textcolor{rebuttle}{To find the largest number of patches the reconstruction attack works, we finetune a pretrained ViT-base to solve the jigsaw puzzle in the settings of $2\times2$, $3\times3$, $4\times4$, and $5\times5$ patches. The training loss converges in the cases of $2\times2$ and $3\times3$ patches, but could not converge in all other cases. This suggests that the threshold is $3\times3$: when the patch number is larger than $3\times3$, the reconstruction attack does not work any more. Our encryption is thus safe from ViT attack, where the patch number is $14\times14$.}
	
	\textbf{Quantitative Measure of Privacy Leakage.}
	\textcolor{rebuttle}{We design a simple measure to quantify the degree of privacy disclosure i.e., detecting faces on original images and encrypted images and then considering the ratio of faces detected on encrypted images as a measure, in which the face detection method used is MTCNN~\footnote{https://github.com/timesler/facenet-pytorch}. We have conducted experiments with this measure, and the results are shown in Table~\ref{tab:leakage}. It can be observed that both RS and MI can protect the visual contents to a great extent.}
	
	\begin{table}[!t]
		\centering
		\caption{\textcolor{rebuttle}{Quantitative measure of privacy leakage. Please note that the experiments are conducted on the validation set of ImageNet.}}
		\label{tab:leakage}
		\begin{tabular}{lcc}
			\hline
			\textcolor{rebuttle}{Image source}          & \textcolor{rebuttle}{\# of Detected faces} & \textcolor{rebuttle}{Ratio of leakage} \\ \hline
			\textcolor{rebuttle}{Original images}       & \textcolor{rebuttle}{15242}                & -                \\
			\textcolor{rebuttle}{Encrypted image by RS} & \textcolor{rebuttle}{1337}                 & \textcolor{rebuttle}{8.8\%}            \\
			\textcolor{rebuttle}{Encrypted image by MI} & \textcolor{rebuttle}{212}                  & \textcolor{rebuttle}{1.4\%}            \\ \hline
		\end{tabular}
	\end{table}
	
	\subsection{Ablation study}
	
\begin{figure*}[!t]
		\centering
		\subfigure{
			\includegraphics[width=0.47\linewidth]{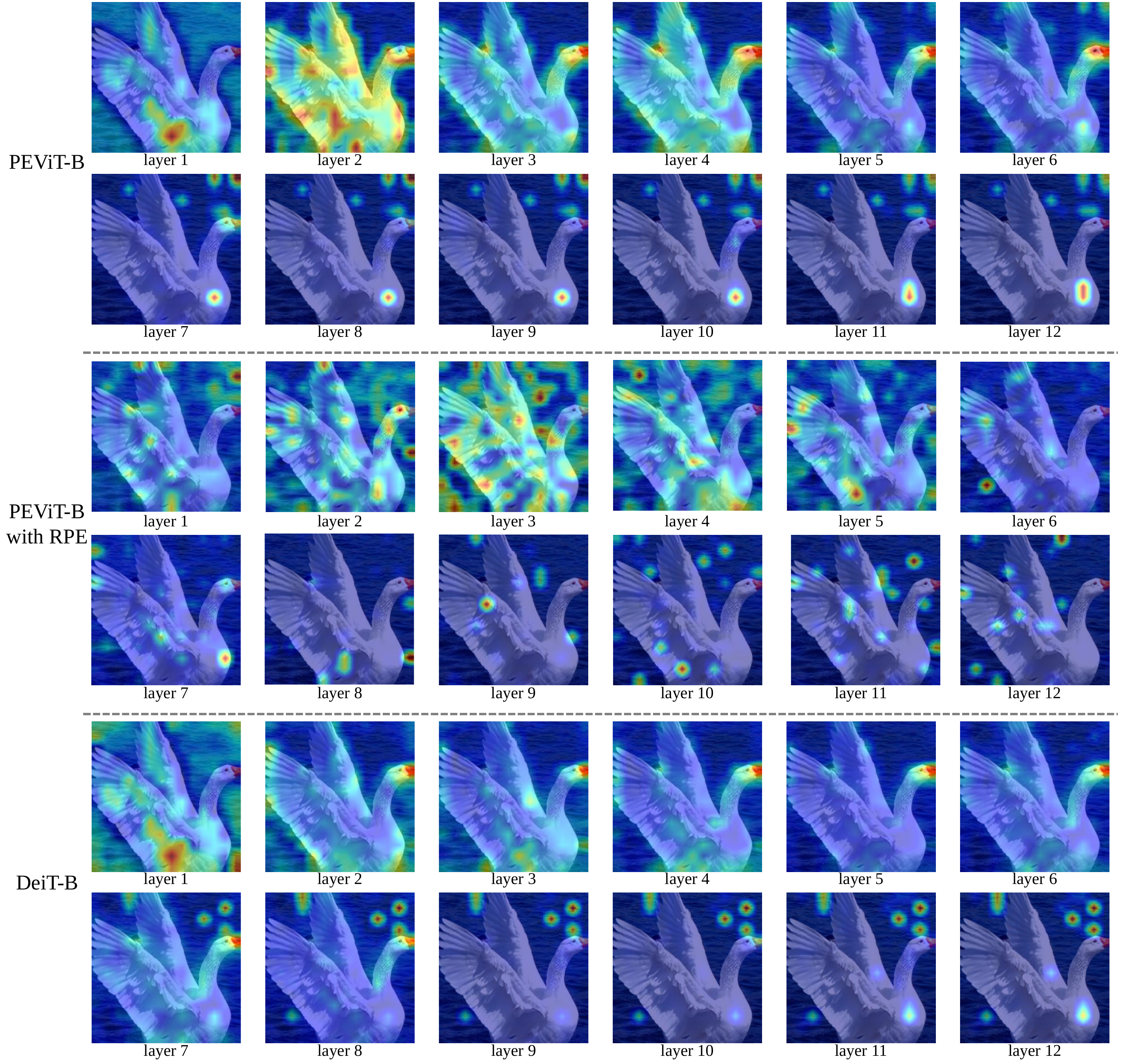}
		}
		\subfigure{
			\includegraphics[width=0.47\linewidth]{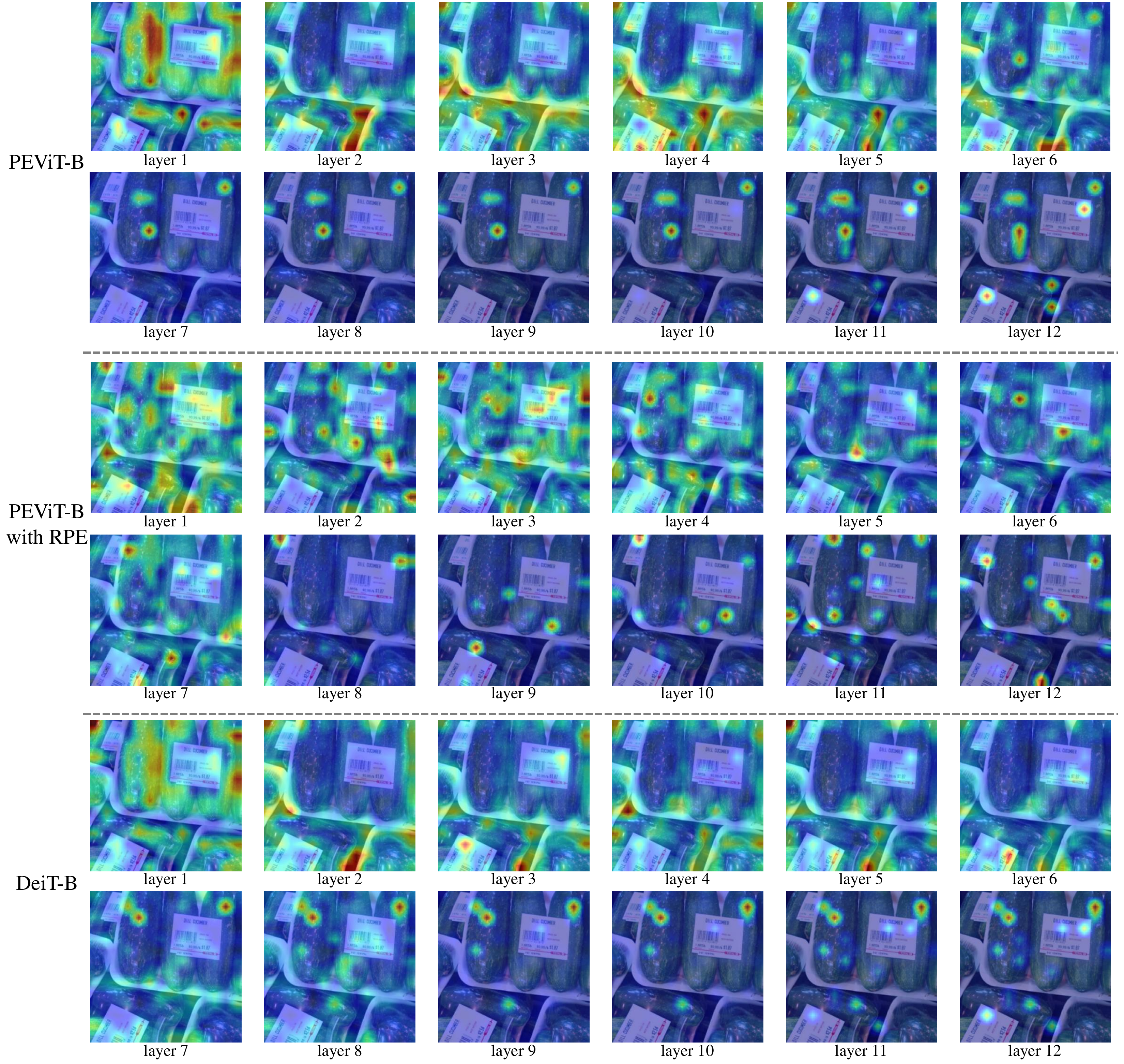}
		}
		\caption{Attention map visualization results of the \emph{class token} in different layers of PEViT-B, PEViT-B with RPE, and DeiT-B for comparison.
		}
		\label{fig:map}
	\end{figure*}

	\textbf{Attention map.}
	We refer to ``discriminative image patches'' as the image patches that contain contents of the target category. To give an intuitive understanding of whether PEViT-B and PEViT-B with RPE can locate discriminative image patches, we provide in Figure~\ref{fig:map} the visualization results of attention maps from different layers of PEViT-B, PEViT-B with RPE, and DeiT-B. 
	It can be observed that (1) The attention of the class token of PEViT-B, PEViT-B with RPE, and DeiT-B becomes more and more concentrated as the number of layers increases and (2) The discriminative image patches of PEViT-B, PEViT-B with RPE, and DeiT-B are only partially attended by the class token at the lower layers. The reason for this might be that the context between image patches plays the key role in the image classification task, or the success of PEViT-B, PEViT-B with RPE, and DeiT-B is largely due to their ability to model high-order co-occurrence statistics between image patches. Therefore, dropping out some patches has no significant impact on performance; see Figure~\ref{fig:attacks}.
	
	\begin{figure*}[!t]
		\centering
		\subfigure[PEViT-B]{
			\includegraphics[width=0.30\linewidth]{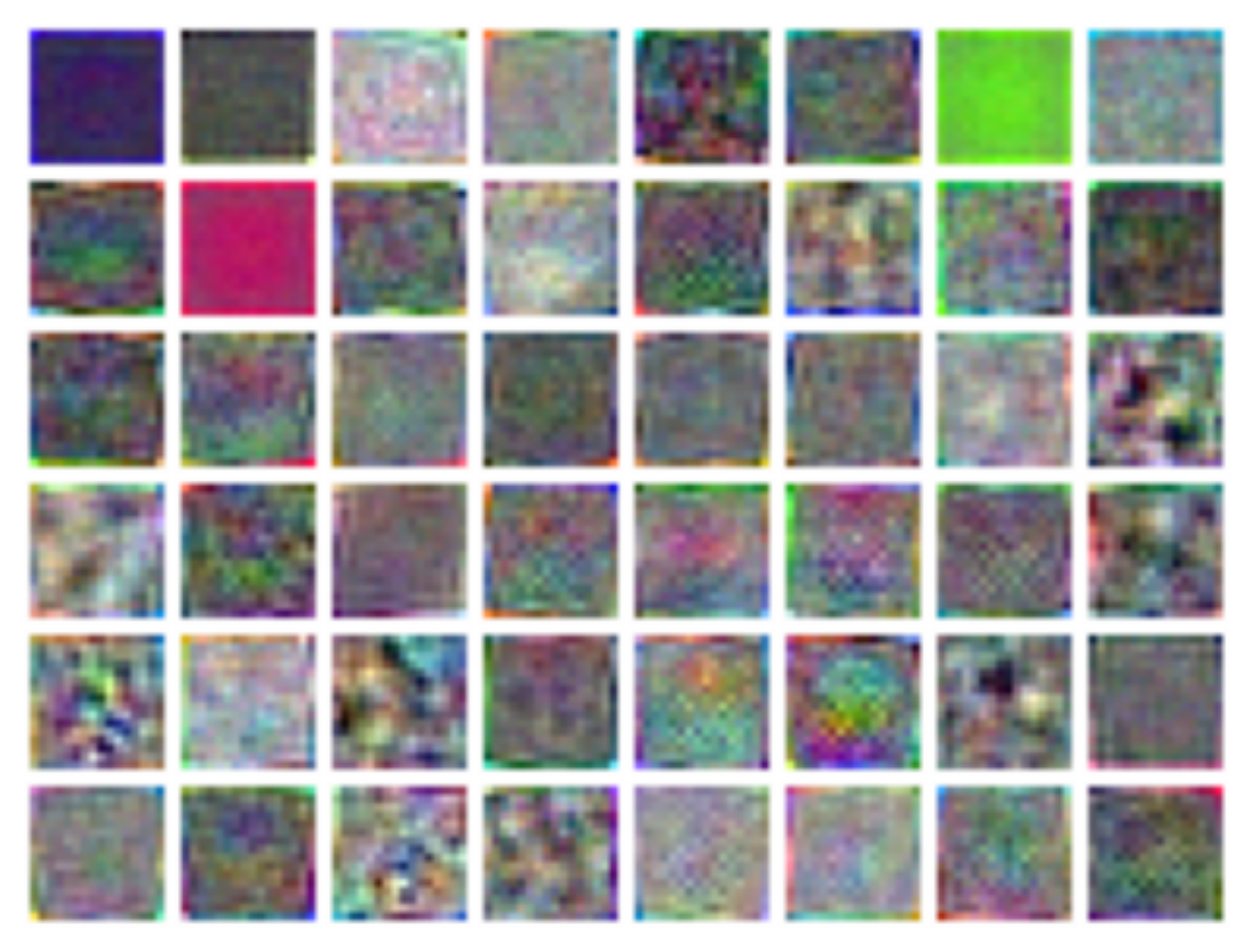}
		}
		\subfigure[PEViT-B with RPE]{
			\includegraphics[width=0.30\linewidth]{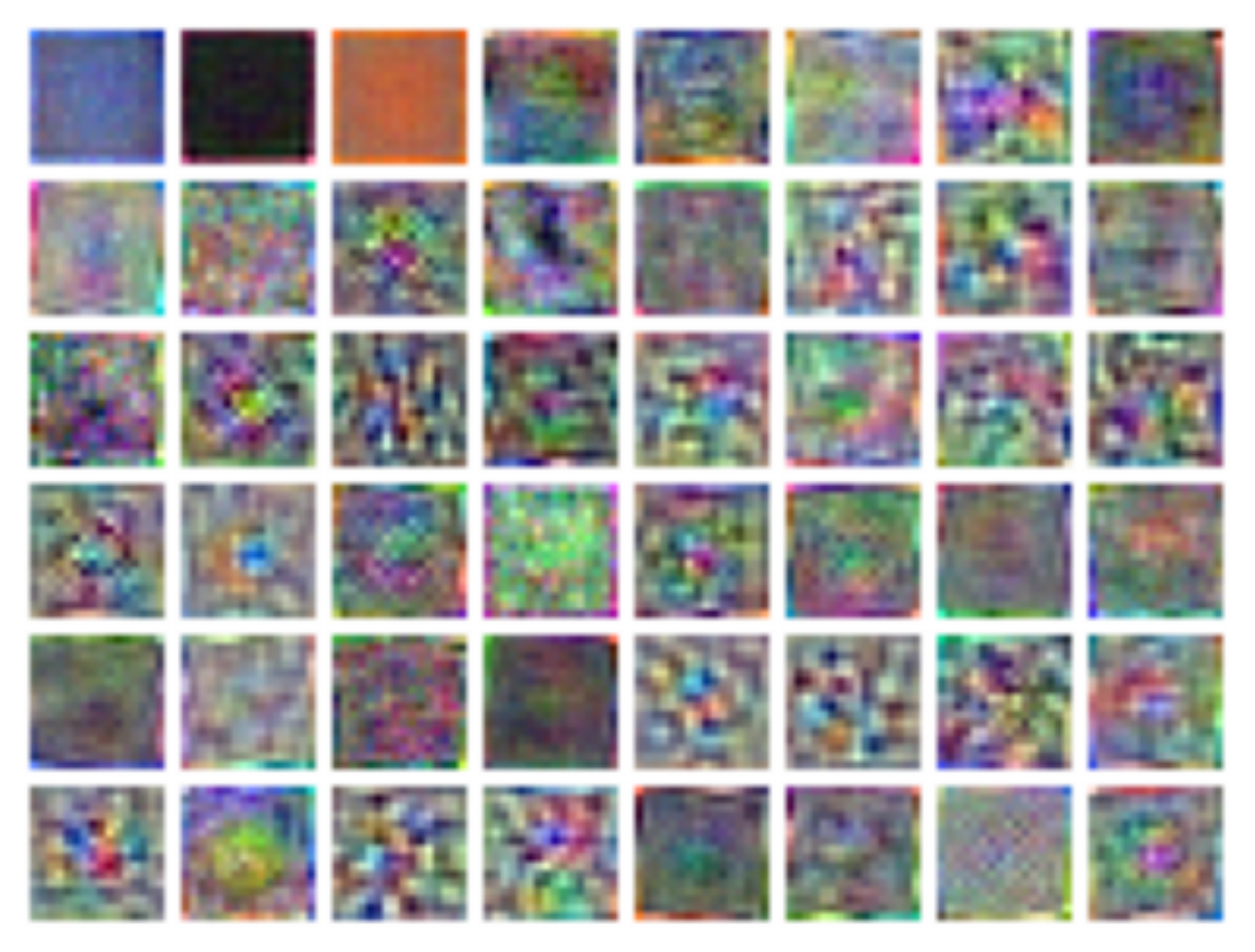}
		}
		\subfigure[DeiT-B]{
			\includegraphics[width=0.30\linewidth]{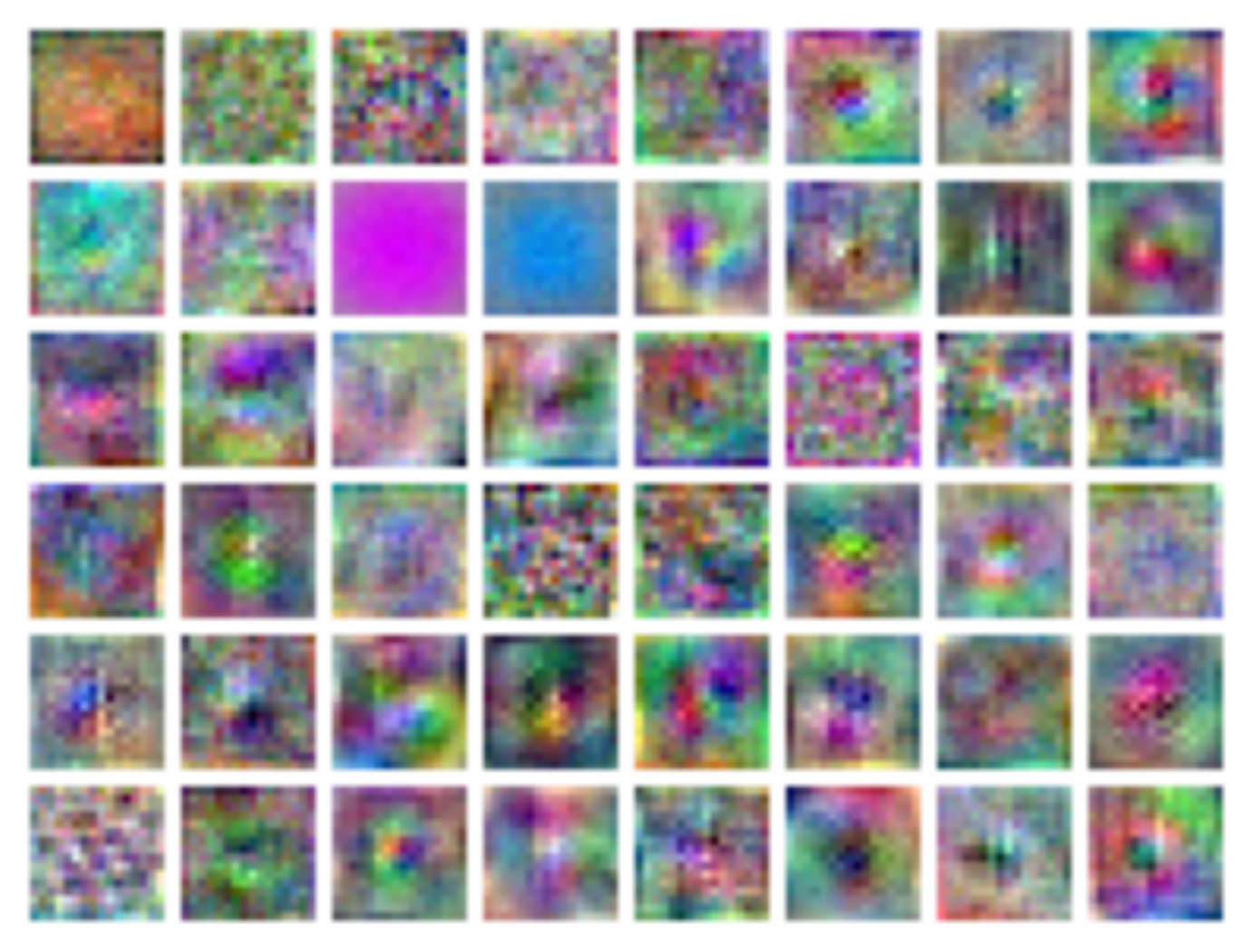}
		}
		\caption{Learned embedding filter visualization. The top 48 principal components of the learned embedding filters are shown.}
		\label{fig:filters}
	\end{figure*}
	
	\begin{figure}[!t]
		\centering
		\includegraphics[width=0.95\linewidth]{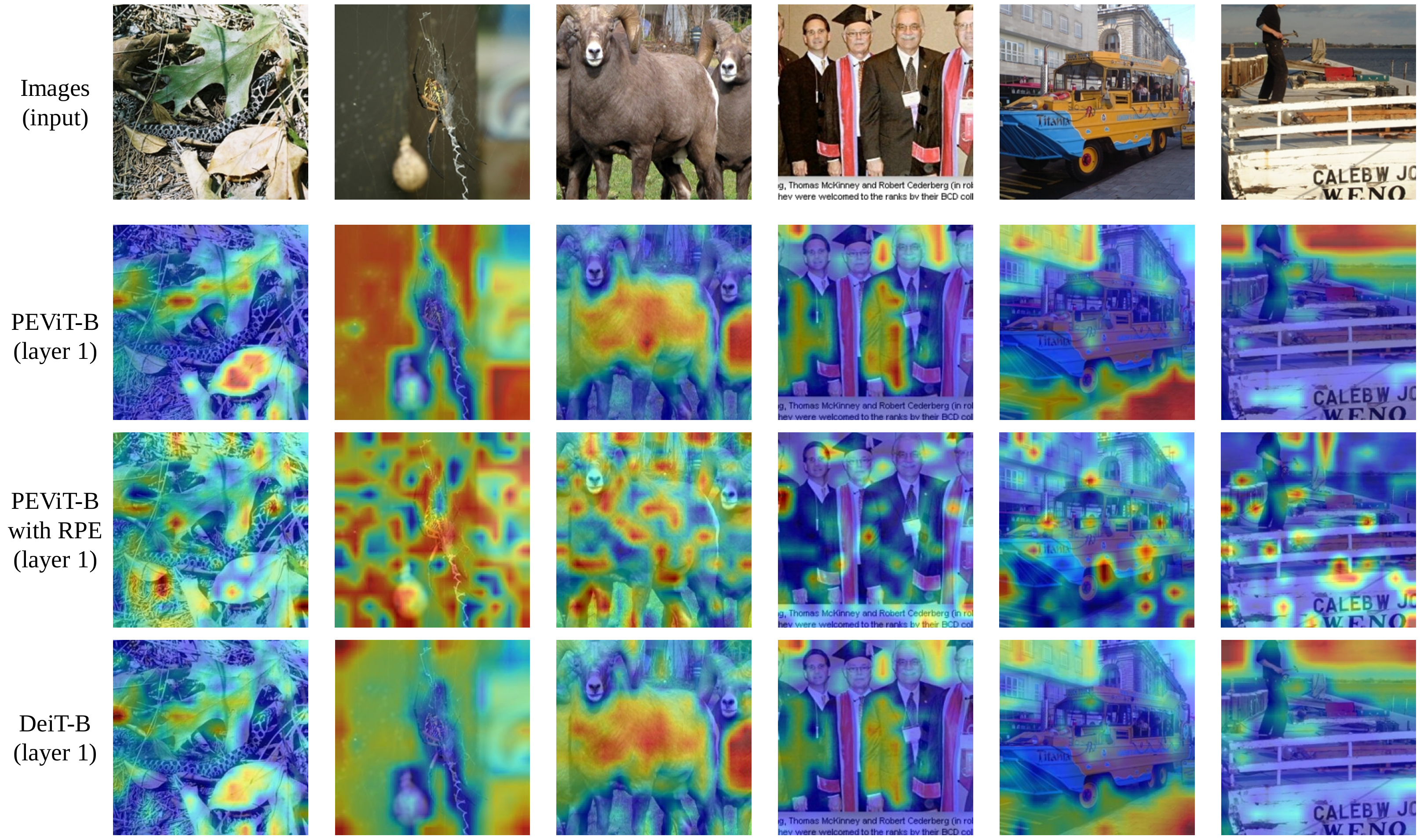}
		\caption{Attention map visualization results from PEViT-B, PEViT-B with RPE, and DeiT-B. The attention maps of the class token in the first layer are shown.
		}
		\label{fig:filters_layer1}
	\end{figure}
	
	\textbf{Learned embedding filter visualization.} To further understand the distinctions between PEViT-B,  PEViT-B with RPE, and DeiT-B, we provide in Figure~\ref{fig:filters} the top 48 principal components of the learned embedding filters. Each filter resembles plausible basis functions for a low-dimensional representation of the fine structure within each patch~\cite{dosovitskiy2020image}. It can be observed that the filters of DeiT-B have more structure than that of both PEViT-B and PEViT-B with RPE. This means that DeiT-B focuses on specific structures that contribute to classification, while PEViT-B and PEViT-B with RPE tend to focus more on the co-occurrence statistics between image patches. Figure~\ref{fig:filters_layer1} shows the attention map of PEViT-B, PEViT-B with RPE, and DeiT-B from the first layer. The distribution of attention of both PEViT-B and PEViT-B with RPE is more dispersed than that of DeiT-B. The reason for this might be that the co-occurrence statistics are more important in PEViT-B and PEViT-B with RPE due to their lack of perception of structural information contained in images. To some extent, this phenomenon is similar to the conclusion in the natural language processing field that higher-order co-occurrence statistics of words play a major role in learning~\cite{wordorderlittle, bagofwords}, as discussed in Sec. 1 of our main paper.
	
	\begin{figure*}[!t]
		\centering
		\subfigure[PEViT-B]{
			\includegraphics[width=0.30\linewidth]{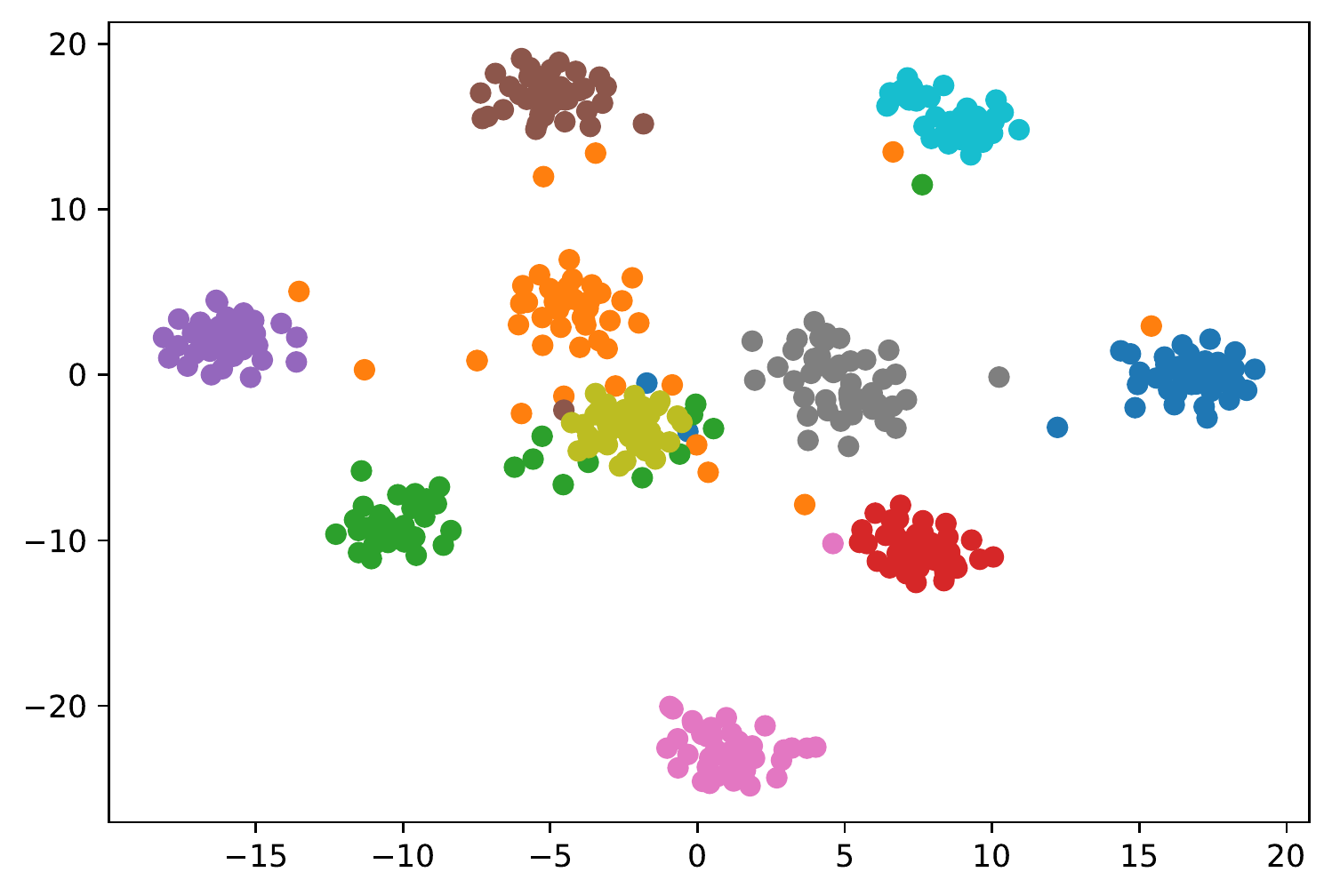}
		}
		\subfigure[PEViT-B with RPE]{
			\includegraphics[width=0.30\linewidth]{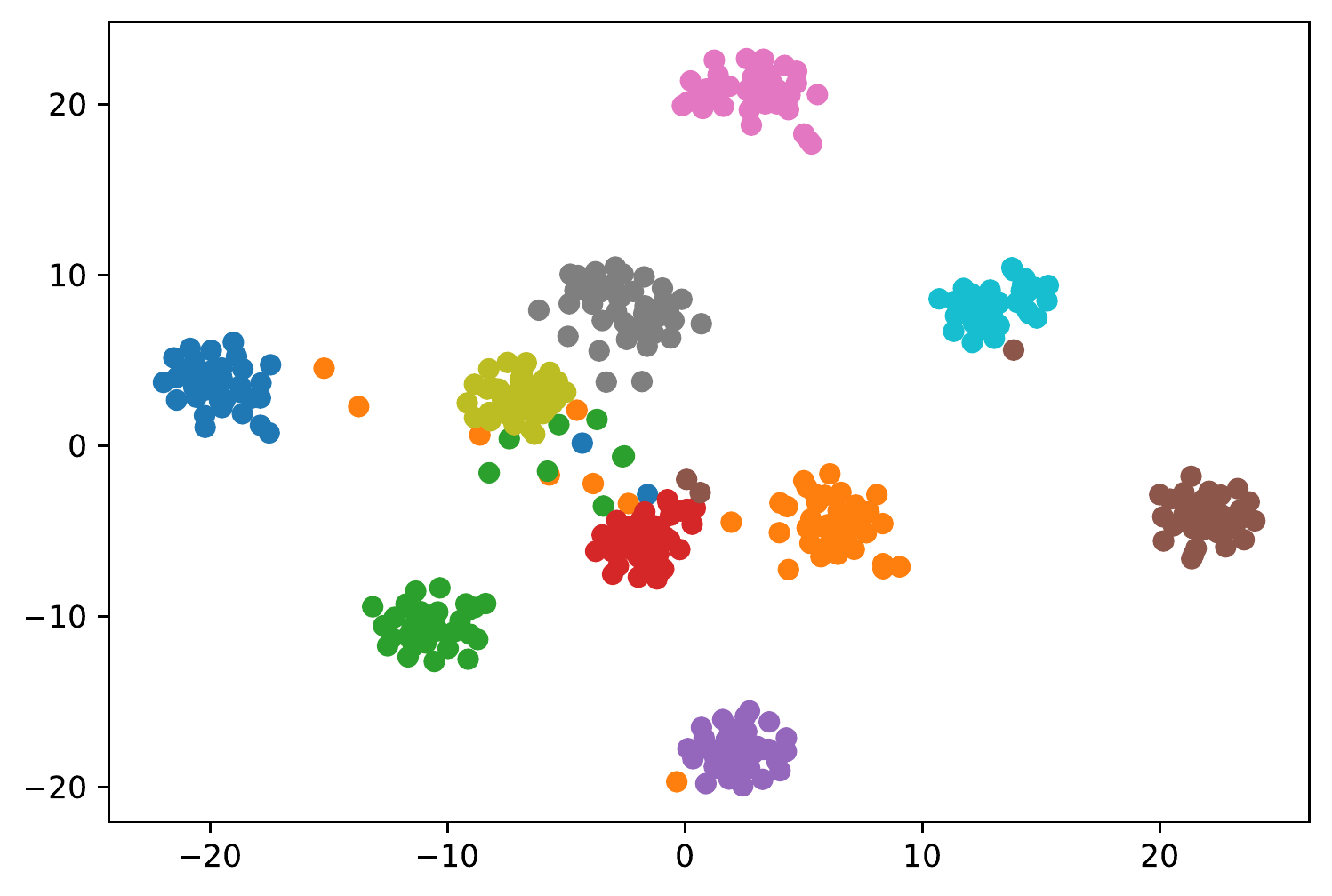}
		}
		\subfigure[DeiT-B]{
			\includegraphics[width=0.30\linewidth]{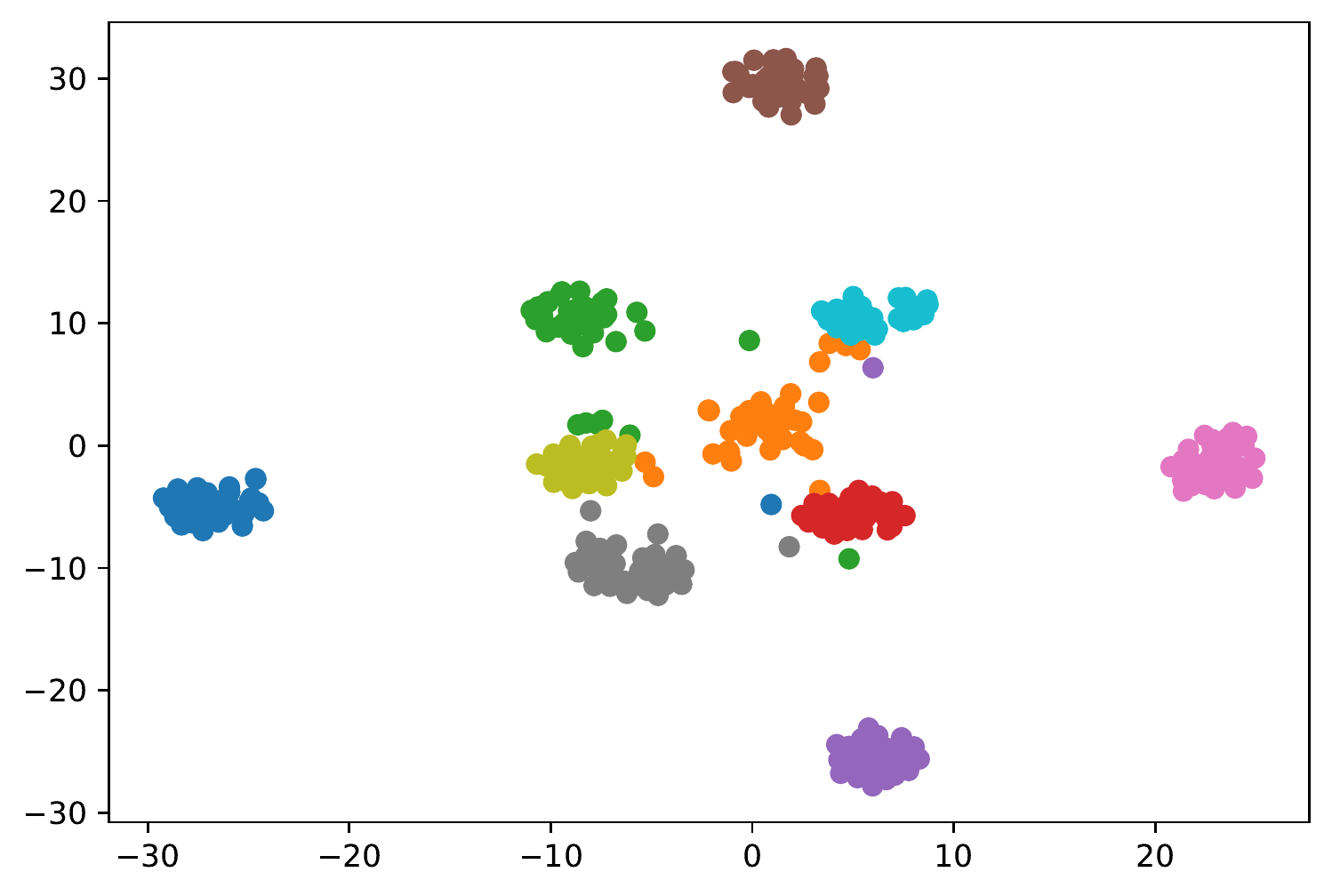}
		}
		\caption{t-SNE feature visualization of PEViT-B, PEViT-B with RPE, and DeiT-B. Features from 10 randomly sampled classes in the validation set are visualized.
		}
		\label{fig:tsne}
	\end{figure*}
	
	\textbf{Feature visualization.} To give an intuitive understanding of the distinctions between PEViT-B, PEViT-B with RPE, and DeiT-B, we provide in Figure~\ref{fig:tsne} the t-SNE feature visualization results of PEViT, PEViT-B with RPE, and DeiT-B. It is worth noting that both PEViT-B and PEViT-B with RPE can be generalized on encrypted images that are randomly shuffled. Interestingly, similar to DeiT-B, our PEViT-B and PEViT-B with RPE can still cluster the features belonging to the same class together while keeping the features belonging to different classes farther apart. This indicates the effectiveness of our permutation-equivariant designs on handling highly encrypted data.
	
	\begin{figure*}[!t]
		\centering
		\subfigure[]{
			\includegraphics[width=0.35\linewidth]{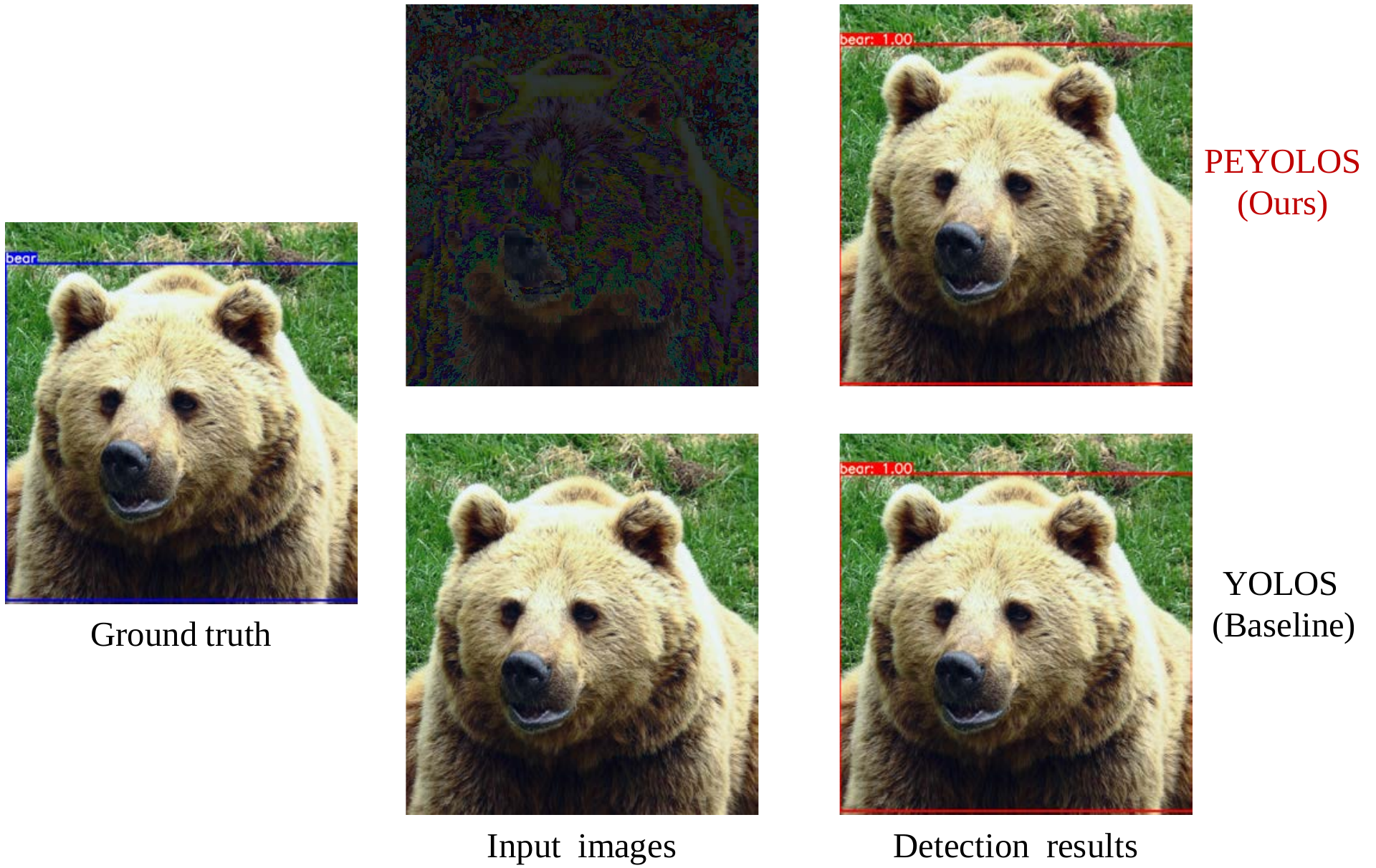}
		}
		\subfigure[]{
			\includegraphics[width=0.5\linewidth]{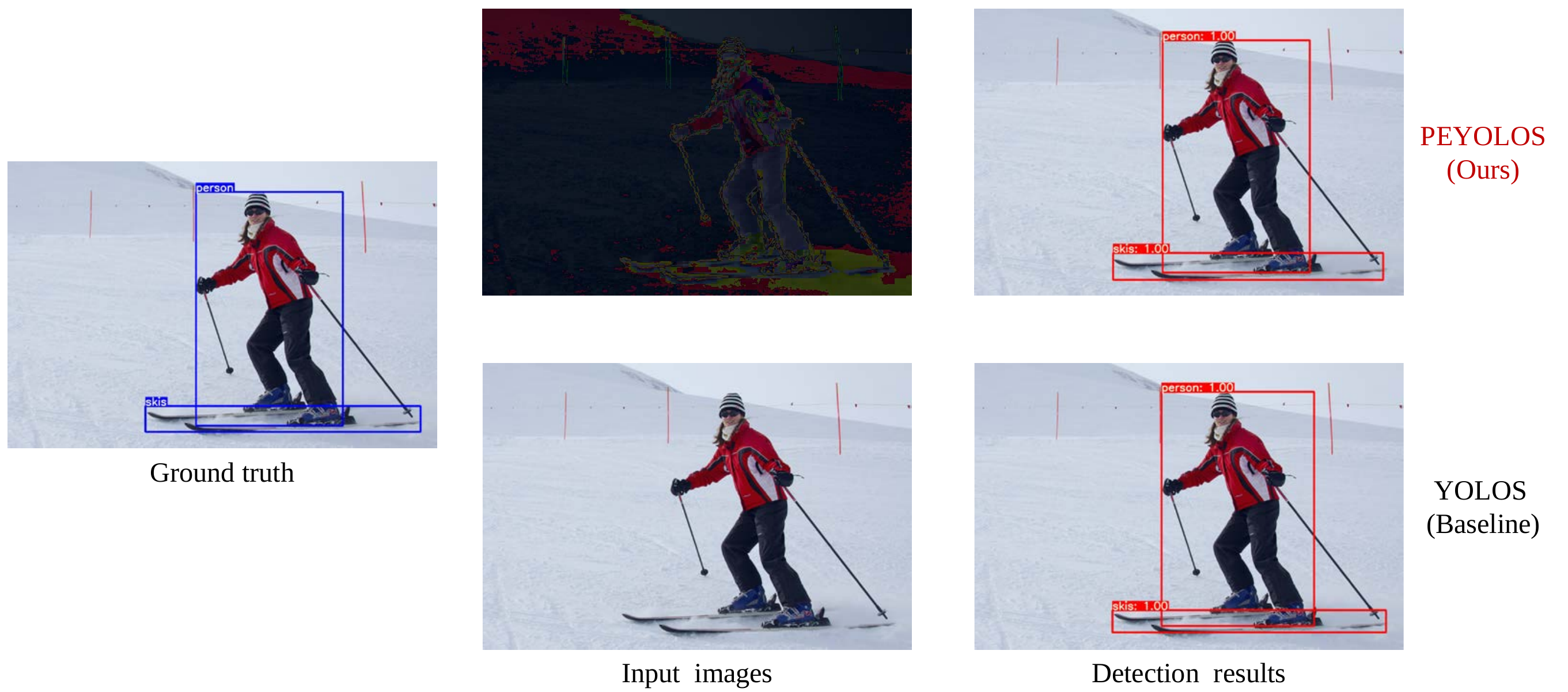}
		}
		\subfigure[]{
			\includegraphics[width=0.45\linewidth]{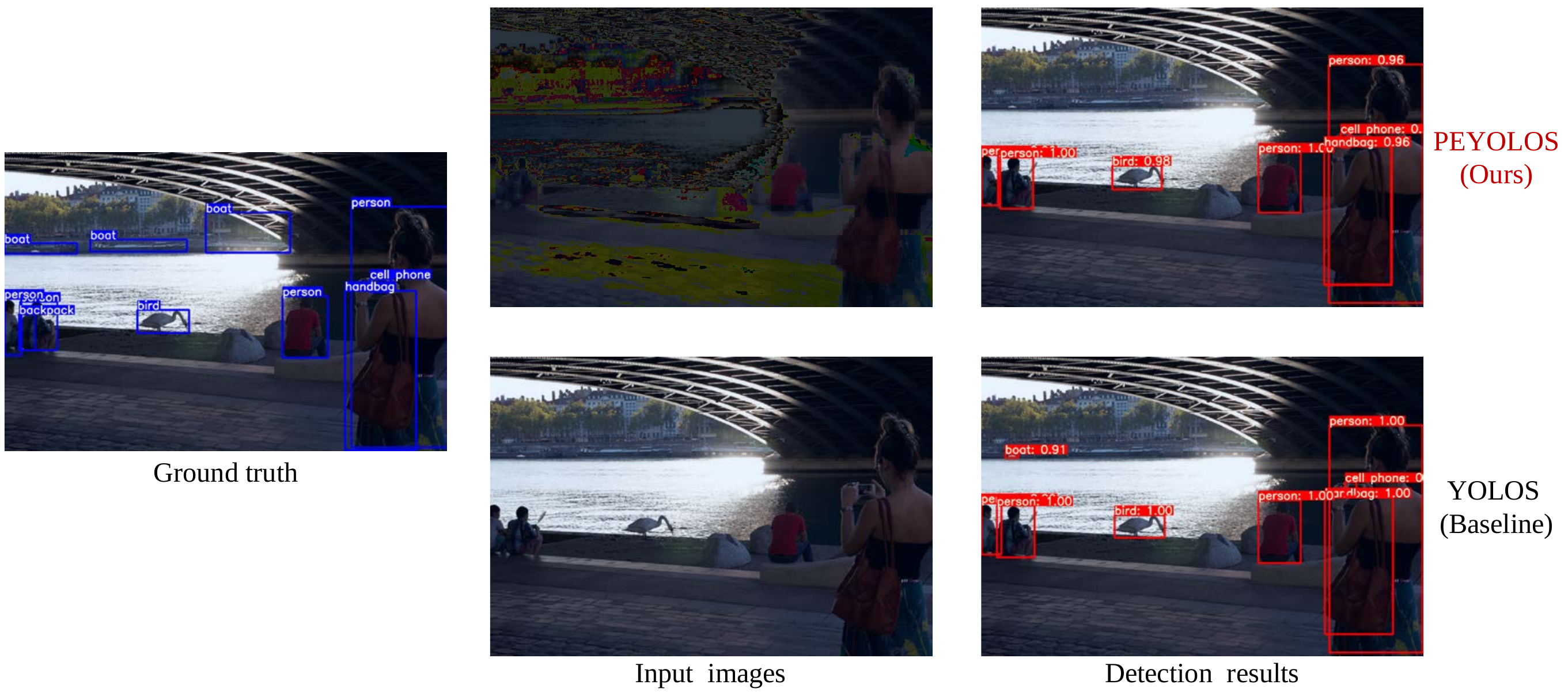}
		}
		\subfigure[]{
			\includegraphics[width=0.45\linewidth]{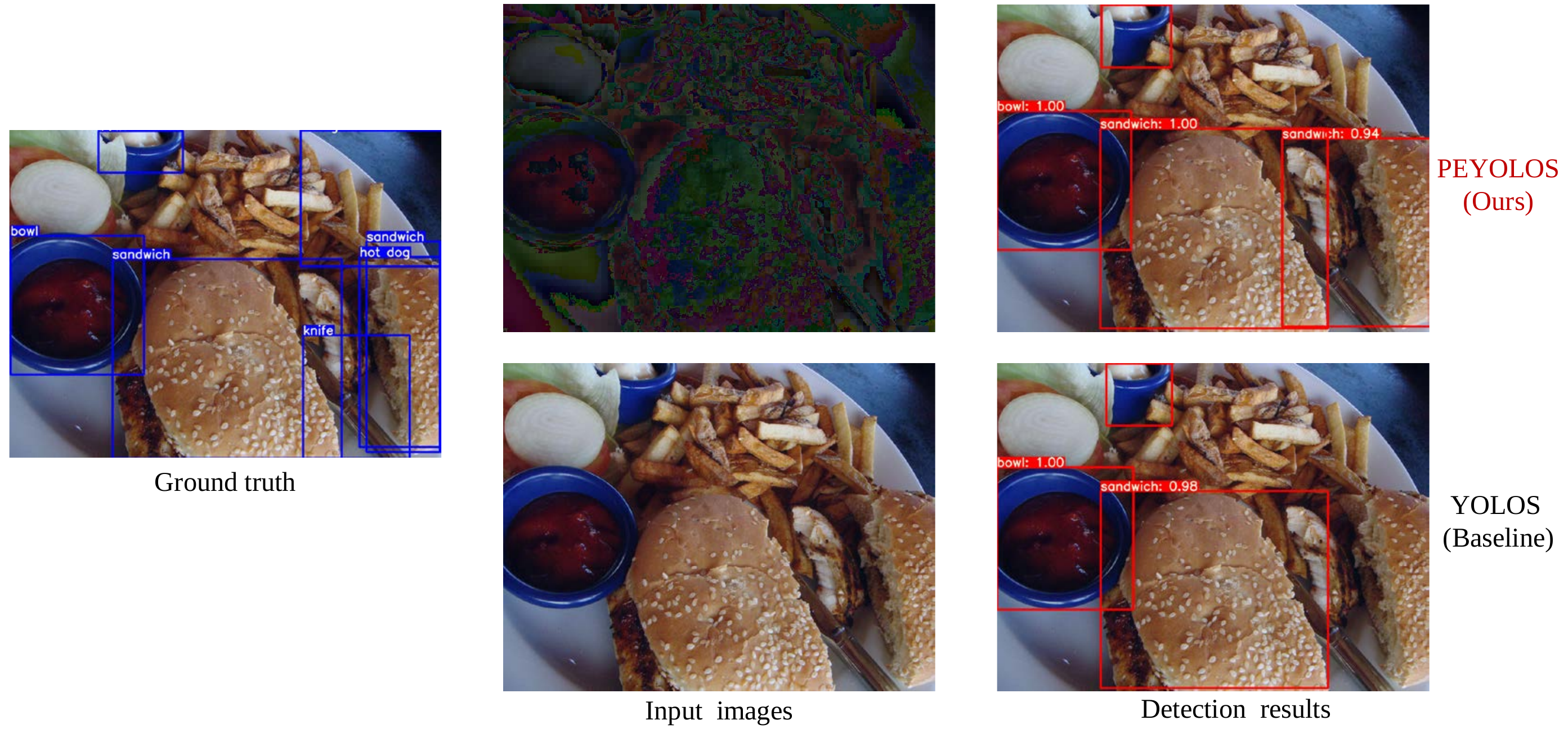}
		}
		\caption{Qualitative results on COCO with the backbone Deit-S. The main difference between PEYOLOS and YOLOS is that PEYOLOS takes highly encryted images which are not human-recognizable and are tough to be decrypted as input while YOLOS take unencrypted images as input.}
		\label{fig:det_results}
	\end{figure*}
	

	\begin{figure*}[!t]
		\centering
		\includegraphics[width=0.90\linewidth]{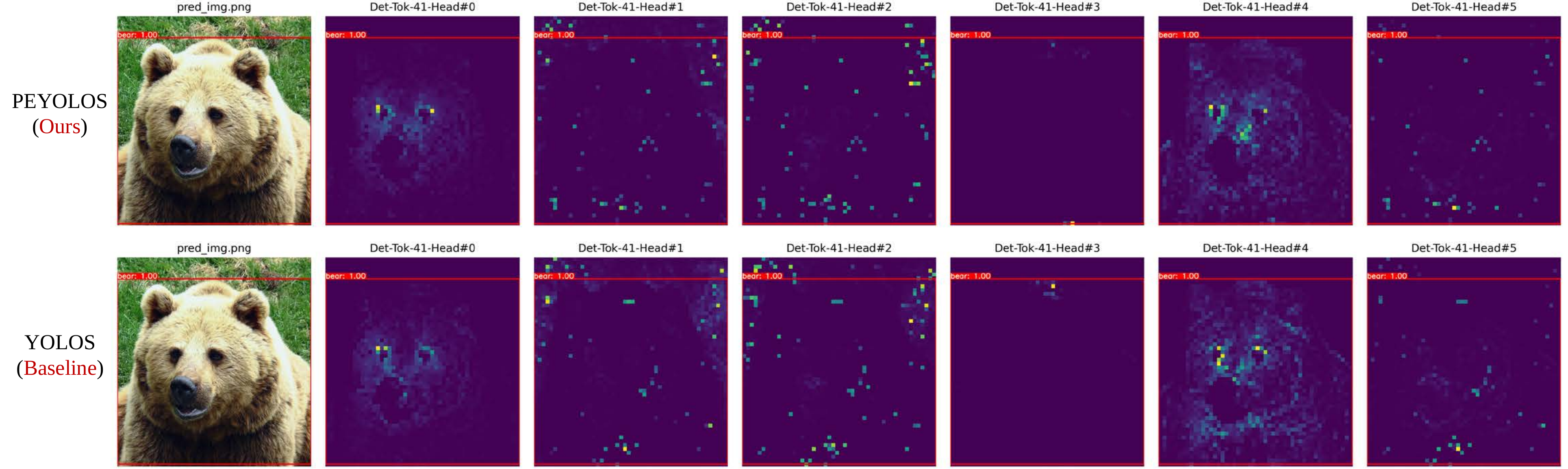}
		\caption{The self-attention map visualization of the detection tokens and the corresponding predictions, of the last layer attention heads based on PEYOLOS (Deit-S) and YOLOS (Deit-S).}
		\label{fig:det_attn_a1}
	\end{figure*}
	
	\textbf{Qualitative results on COCO.}
	To show the difference between PEYOLOS and YOLOS, we provide in Figure~\ref{fig:det_results} the qualitative results on COCO of both PEYOLOS and YOLOS. The main difference between PEYOLOS and YOLOS is that PEYOLOS takes encrypted images as input while YOLOS take unencrypted images as input. Compared with YOLOS, the performance degradation of PEYOLOS is only $\sim 3\%$. Based on these observations, we conclude that the proposed learning paradigm can destroy human-recognizable contents while preserving machine-learnable information.
	
	\textbf{Attention maps of detection tokens.}
	We inspect the attention maps of detection tokens that are related to the predictions. The experiments are conducted based on PEYOLOS and YOLOS, and the visualization results of the last layer are shown in 
	Figure~\ref{fig:det_attn_a1}. It can be observed that (1) For both PEYOLOS and YOLOS, different attention heads focus on different patterns at different locations, (2) The corresponding detection tokens as well as the attention map patterns are usually different for PEYOLOS and YOLOS, and (3) Some visualizations are interpretable while others are not. The reason for this might be that both image patches and their context plays the key role in the object detection task and how the high-order co-occurrence statistics between image patches affect the detection results still remains unclear.
	
	\begin{table}[!t]
		\centering
		\caption{The impact of patch size on object detection performance. Here, the backbone used is DeiT-Ti.}
		\label{tab:patchsize}
		\begin{tabular}{lc}
			\hline
			\textcolor{rebuttle}{Patch size}          & \textcolor{rebuttle}{Accuracy (AP)} \\ \hline
			\textcolor{rebuttle}{$20\times20$}        & \textcolor{rebuttle}{20.7}          \\
			\textcolor{rebuttle}{$16\times16$}        & \textcolor{rebuttle}{25.3}          \\
			\textcolor{rebuttle}{$14\times14$}        & \textcolor{rebuttle}{23.4}          \\
			\textcolor{rebuttle}{$10\times10$}        & \textcolor{rebuttle}{23.2}          \\ \hline
		\end{tabular}
	\end{table}
	
	\textbf{Ablation on Patch Size for Object Detection.}
	\textcolor{rebuttle}{We have conducted ablation experiments on the patch size for object detection, in which the backbone used is DeiT-Ti. The results are shown in Table~\ref{tab:patchsize}. It can be observed that the best performance is achieved with the patch size of $16\times16$. Therefore, the patch size is set to $16\times16$ by default.}

	\section{Conclusion}
	
	In this paper, we propose an efficient privacy-preserving learning paradigm that can destroy human-recognizable contents while preserving machine-learnable information. The key insight of our paradigm is to decouple the encryption algorithm from the network optimization via permutation-invariance. Two encryption strategies are proposed to encrypt images: random shuffling to a set of equally-sized image patches and mixing image patches that are permutation-invariant. By adapting ViT and YOLOS with minimal adaptations, they can be made (partially) permutation-invariant and are able to handle encrypted images. Extensive experiments on ImageNet and COCO show that the proposed paradigm achieves comparable accuracy with the competitive methods, meanwhile destroying human-recognizable contents.
	
	\section{Data Availability Statements} 
	The datasets used in this study (i.e., \href{https://www.image-net.org/}{ImageNet} and \href{https://cocodataset.org/#home}{COCO}) are all publicly available for the research purpose.
	
	\bibliography{iclr2023_conference}
	
\end{document}